\documentclass{article} % For LaTeX2e
\usepackage[final]{nips_2016}
\usepackage{hyperref}
\usepackage{natbib}
\usepackage{amsmath}
\usepackage{url}
\usepackage{lipsum}
\usepackage{fancyhdr}
\usepackage{datetime}

\usepackage{graphicx}
\usepackage{caption}

\usepackage{booktabs}
\usepackage{tabularx}
\usepackage{multirow}
\usepackage{datetime}
\usepackage{relsize}
\usepackage{amsmath,amssymb,textcomp}
\usepackage{nicefrac}
\usepackage{amsthm}
\usepackage{tikz}
\usepackage{nth}
\usepackage{rotating}
\usepackage{array}
\usepackage{fp}
\usepackage{stringstrings}
\usepackage{pifont}
\usepackage[capitalise]{cleveref}

\newcommand{\define}{\triangleq}

\newcommand{\layerindex}{\ell}
\newcommand{\vect}[1]{{\bf{#1}}}
\newcommand{\mat}[1]{\mathbf{#1}}
\newcommand{\R}{\mathbb{R}}
\newcommand{\cL}{\mathcal{L}}
\newcommand{\Bb}{\mathbf{b}}

\newcommand{\vb}{\vect{b}}

\newcommand{\vO}{\mat{O}}

\newcommand{\vW}{\mat{W}}

\newcommand{\vX}{\mat{X}}

\newcommand{\vy}{\vect{y}}

\newcommand{\x}{\mathbf{x}}
\newcommand{\y}{\mathbf{y}}

\newcounter{mnote}

\usepackage{subcaption}
\usepackage{tablefootnote}
\title{DropNeuron: Simplifying the Structure of Deep Neural Networks}

\author{
Wei Pan \\
Data Science Institute \\
Imperial College London \\
w.pan11@imperial.ac.uk
\And 
Hao Dong \\
Data Science Institute\\
Imperial College London\\
hd311@imperial.ac.uk
\And 
Yike Guo \\
Data Science Institute\\
Imperial College London\\
y.guo@imperial.ac.uk
}

\newcommand{\code}{\url{http://www.github.com/panweihit/DropNeuron}}

\begin{document}
 
\maketitle
\begin{abstract}

Deep learning using multi-layer neural networks (NNs) architecture manifests superb power in modern machine learning systems. The trained Deep Neural Networks (DNNs) are typically large.  The question we would like to address is whether it is possible to simplify the NN during training process to achieve a reasonable performance within an acceptable computational time. We presented a novel approach of optimising a deep neural network through regularisation of network architecture. We proposed regularisers which support a simple mechanism of dropping neurons during a network training process. The method supports the construction of a simpler deep neural networks with compatible performance with its simplified version. As a proof of concept, we evaluate the proposed method with examples including sparse linear regression, deep autoencoder and convolutional neural network. The valuations demonstrate excellent performance.

The code for this work can be found in \code

\end{abstract}

\section{Introduction}

It is commonly accepted for a deep learning system, the underlying neural network (NN) has to be big and complex. We argue that this perception may not be true. Many of the neurons and their associated connections, both incoming and outgoing ones, can be \emph{dropped permanently} which results in a NN with much smaller size. This is very similar to sparse distributed representations in brain. The human neocortex has roughly 100 billion neurons, but at any given time only a small percent are active in performing a particular cognitive function \citep{olshausen1997sparse}.  For the non-sequence or non time dependent data, the active neurons may be fixed and not change over time \citep{cui2016htm}.  Dropping neurons is also the key idea in Dropout \citep{hinton2012improving,srivastava2014dropout}, a successful regularisation technique to prevent overfitting in NNs. In their work, the neurons are \emph{dropped temporarily in training}. In the end for prediction, the model is still of full size and fully connected. 

Hereafter, we aim at training a simple network when it can achieve comparable performance to the fully connected NN, but with number of neurons and connections as few as possible.
Dropping connections may be not difficult by introducing weight decay regularisers. However, dropping neurons is challenging. On one hand, the weight decay regularisation can't penalise \emph{all} the connections associated with one neuron simultaneously. On the other hand, it is attempted to suppress the neurons to fire such as the use of rectifier as activation function \citep{glorot2011deep}, regularisation techniques like K-L sparsity in the sparse autoencoder variants \citep{kingma2013auto, bengio2013representation}, or constraints like max-norm \citep{srebro2005rank, goodfellow2013maxout}. However, a neuron not firing in training still can't be dropped for testing and prediction since her connections' weights are not zeros.
As an alternative, network pruning by dropping connections below a threshold has been widely studied to compress a pre-trained fully connected NN models reduce the network complexity and over-fitting, see early work \citep{lecun1989optimal, hassibi1993second} and more recently \citep{han2015learning, han2015deep_compression}. Unfortunately, such pruning strategy may not effectively drop neurons. For example, a NN may consist of large number of neurons but few connections. Though, the model size/storage space may not be challenging but brings another challenge for chip design for storage and computation, e.g. (mobile) GPU, FPGA, etc. For example, sparse matrix computation still

In this paper, we propose a strategy to drop neurons. A neuron can be dropped by regularising all her incoming connections' weights and/or all her outgoing connections' weights to other neurons to be \emph{zeros}. 
Furthermore, we will show that to achieve a simplest network is intractable but some convex relaxation over the cost function can alleviate the difficulty. Such relaxation may yield a simple network yet (maybe) not minimal.
It can be realised by introducing two new regularisers to penalise incoming and outgoing connections respectively. Both regularisers have a form $\Vert \cdot \Vert_2 $ which is inspired by Group Lasso \citep{yuan2007model}.
In the end, we test our strategy by three tasks: the first one is on sparse linear regression which is widely used as benchmark in Compressive Sensing \citep{candes2005decoding, donoho2006compressed}; the second one is on unsupervised learning using Autoencoder for MNIST data; the third one is to use convolutional NN with LeNet-5 structure for classification of MNIST data. The evaluation demonstrates the possibility of dropping neurons but still achieving good performance.

 \section{Dropping Neurons of Deep Multilayer Perceptron Architecture}

We use the following notation throughout the paper. Bold lower case letters ($\x$) denote vectors, bold upper case letters ($\vX$) denote matrices, and standard weight letters ($x$) denote scalar quantities. 
We use subscripts to denote variables as well $\vW^\layerindex$ (such as $\vW^1: {n^{0} \times n^{1}}, \vW^2: {n^{1} \times n^{2}}$). $n^{0}$ is the number of features of the input.
We use subscripts to denote either entire rows ($\vW^\layerindex_{p,:}$ for the $p$-th row of $\vW^\layerindex$) or entire columns ($\vW^\layerindex_{:,q}$ for the $q$-th column of $\vW^\layerindex$). 
We use the standard capital letter with subscript to denote the element index of a specific variable: $W^1_{p,q}$ denotes the element at row $p$ column $q$ of the variable $\vW^1$.
We also use $\vO^\layerindex$ to be the indicator for the neurons in layer $\layerindex$. For example, $\vO^0$ consist of $n^{0}$ neurons in the input layer, indexed as $O^0_1, \ldots, O^0_{n^{0}}$.

\subsection{Regularisation} \label{sec:cost}
We start with the case of a three layer NN with \textit{a single hidden layer}. 
The generalisation to multiple layers is straightforward. Denote by $\vW^1, \vW^2$ the weight matrices connecting the first layer to the hidden layer and connecting the hidden layer to the output layer respectively. These linearly transform the layers' inputs before applying some element-wise non-linearity $\sigma(\cdot)$.
Denote by $\Bb$ the biases by which we shift the input of the non-linearity. 
We assume the model to output $n^2$ dimensional vectors while its input is $n^0$ dimensional vectors, with $K$ hidden units. Thus $\vW^1$ is a ${n^{0} \times n^{1}}$ matrix, $\vW^2$ is a ${n^{1} \times n^{2}}$ matrix, and $\Bb$ is a $n^1$ dimensional vector. A standard NN model would output the following given some input $\x$
\begin{equation}
\begin{aligned}
\widehat{\y} = \sigma(\x \vW^1 + \Bb) \vW^2
\label{model:output}
\end{aligned}
\end{equation}

To use the NN model for regression we might use the Euclidean loss (also known as ``square loss''),
\begin{align} \label{eq:NN:reg}
E_{\text{regression}} = \frac{1}{2N}
\sum_{n=1}^N ||\y_n - \widehat{\y}_n||^2_2
\end{align}
where $\{\y_1, \hdots, \y_N\}$ are $N$ observed outputs, and $\{\widehat{\y}_1, \hdots, \widehat{\y}_N\}$ being the outputs of the model with corresponding observed inputs $\{ \x_1, \hdots, \x_N \}$. 

To use the model for classification, predicting the probability of $\x$ being classified with label $1,...,D$, we pass the output of the model $\widehat{\y}$ through an element-wise softmax function to obtain normalised scores: $\widehat{p}_{nd} = \exp(\widehat{y}_{nd}) / \left(\sum_{d'} \exp(\widehat{y}_{nd'})\right)$. Taking the log of this function results in a \textit{softmax} loss,
\begin{align} \label{eq:NN:class}
E_{\text{classification}} = -\frac{1}{N} \sum\limits_{n=1}^N \log(\widehat{p}_{n,c_n})
\end{align}
where $c_n \in [1, 2, ..., D]$ is the observed class for input $n$.

During optimisation regularisation terms are often added. Some of the well known regularisation include $\ell_1$ regularisation and $\ell_2$ regularisation, defined as
\begin{equation}
\begin{aligned}
\texttt{l1$\_$regulariser} := \lambda_{\ell_1} \sum_{\layerindex=1}^{L} (\Vert\vW^\layerindex\Vert_1+\Vert\vb^\layerindex\Vert_1)
\  \text{and} \
\texttt{l2$\_$regulariser} := \lambda_{\ell_2} \sum_{\layerindex=1}^{L} (\Vert\vW^\layerindex\Vert_2^2+\Vert\vb^\layerindex\Vert_2^2)
\end{aligned}
\end{equation}
where $\lambda_{\ell_1}$ and $\lambda_{\ell_2}$ are often called weight decay or regularisation parameter which needs fined tuned.
 
Then it results in a minimisation objective (often referred to as cost),
\begin{equation}
\begin{aligned}
\cL := E + \texttt{l1$\_$regulariser} \ \ \ \ \text{or} \ \ \ \ \cL := E + \texttt{l2$\_$regulariser}, 
\label{cost:l1}
\end{aligned}
\end{equation}
or a mixture of  \texttt{l1$\_$regulariser} and  \texttt{l2$\_$regulariser}, which is known as elastic net.

The goal of introducing \texttt{l1$\_$regulariser} and \texttt{l2$\_$regulariser} is to penalise the connections' weights between neurons to prevent overfitting. However, the application of such regularisers alone in deep neural network are not as successful as in linear regression and logistic regression.
On the other hand, in the hardware computation especially using GPU, dropping connections may not save computation time and memory unless some special coding and processing is used \citep{han2015deep_compression}. The introduction of dropout achieve great success to avoid over-fitting in practice \citep{hinton2012improving,srivastava2014dropout} with these two regularisers. 
These regularisation techniques are suitable for preventing overfitting but may not be helpful in simplifying the NN structure. We believe that the key to automatically simplify a NN structure in training is to define proper regulariser by exploring the sparsity structure of the NN in a deep learning system.

\subsection{Dropping Neurons by Regularisation}
\label{sec:strategy}

Hereafter, we are seeking a strategy to drop neurons. Using the standard setup for NN,  we have the weight matrix from layer $\layerindex-1$ to layer $\layerindex$, 
\begin{equation}
\begin{aligned}
\vW^\layerindex =\left[ (\vW^\layerindex_{1,:})^{\top}, \ldots, (\vW^\layerindex_{n^{\layerindex-1},:})^{\top} \right]^{\top} = \left[ \vW^\layerindex_{:,1}, \ldots, \vW^\layerindex_{:, n^{\layerindex}}\right]
\label{}
\end{aligned}
\end{equation}
where $\vW^\layerindex_{i,:}$ denote the $i$-th row of $\vW^\layerindex$, $i = 1, \ldots,n^{\layerindex-1}$; it encodes the incoming connections' weights from layer $\layerindex-1$ to the $i$-th neuron in layer $\layerindex $, i.e., $O^\layerindex_i$.
Similarly, $\vW^\layerindex_{:,j}$ denote the $j$-th column of $\vW^\layerindex$, $j = 1, \ldots,n^{\layerindex}$; it encodes the outgoing connections' weights of the $j$-th neuron in layer $\layerindex $, i.e., $O^\layerindex_i$ to all the neurons in the next layer, i.e., layer $\layerindex+1$ . In particular, $O^0_i$ denotes the $i$-th feature/neuron in input layer.

\subsubsection{New Regularisers}
We first introduce two new regularisers, the first one is called \texttt{li$\_$regulariser}$(\lambda_{\ell_i})${\footnote{\texttt{i} in \texttt{li$\_$regulariser} denotes the initials of \underline{i}n-coming which resembles the column removal in Fig. \ref{fig:removecolumn}}}
\begin{equation}
\begin{aligned}
\texttt{li$\_$regulariser} :=\lambda_{\ell_i} \sum_{\layerindex=1}^{L}\sum_{j=1}^{n^{\layerindex}}  { \Vert  \vW^\layerindex_{:,j} \Vert_2}
=
\lambda_{\ell_i} \sum_{\layerindex=1}^{L}\sum_{j=1}^{n^{\layerindex}}{ \sqrt{\sum_{i=1}^{n^{\layerindex-1}} \left(W_{ij}^{\layerindex}\right)^2}}
\label{li_regularizer}
\end{aligned}
\end{equation}
This is used to regularise the incoming connections' weights of all the neurons across different layers over the whole network.

The second one is called \texttt{lo$\_$regulariser}$(\lambda_{\ell_o})$\footnote{\texttt{o} in \texttt{lo$\_$regulariser} denotes the initials of \underline{o}ut-going which resembles the row removal in Fig. \ref{fig:removerow}}
\begin{equation}
\begin{aligned}
\texttt{lo$\_$regulariser} :=\lambda_{\ell_o}\sum_{\layerindex=1}^{L} \sum_{i=1}^{n^{\layerindex-1}}\Vert  \vW^\layerindex_{i,:} \Vert_2
=
\lambda_{\ell_o}\sum_{\layerindex=1}^{L} \sum_{i=1}^{n^{\layerindex-1}}{ \sqrt{\sum_{j=1}^{n^{\layerindex}} \left(W_{ij}^{\layerindex}\right)^2}}
\label{lo_regularizer}
\end{aligned}
\end{equation}
This is used to regularise the outgoing connections' weights of all the neurons across different layers over the whole network. The key idea of introducing the two regularisers is to embed a dropping mechanism in a deep NN training process. Such a dropping mechanism is guided by the two regularisers.

\subsubsection{Dropping Principles}

\paragraph{Dropping Principle 1} 
First of all, we perform network pruning for small weights. After training, some of the estimated weights tend to be (very) small, e.g. to the magnitude less than $10^{-3}$. A straightforward idea is to prune/remove weights below a threshold to reduce the network complexity and over-fitting. Actually the idea on network pruning is not new and proved to be a valid \citep{lecun1989optimal}. 
And recently \citet{han2015learning, han2015deep_compression} pruned state-of-the-art fully connected NN models which are pre-trained with no loss of accuracy. The key is to select a proper threshold to drop connections. Our dropping principle is similar to the previous work and fairly simple: no loss of accuracy after pruning.  As shown in \citep{han2015learning, han2015deep_compression} and our experiments, pruning reduced the number of parameters by over $10\times$. Unfortunately, such pruning can not effectively drop neurons.

\begin{figure} %[h]
	\centering
	\begin{subfigure}[b]{0.47\textwidth}
		\centering
		\includegraphics[width=\textwidth]{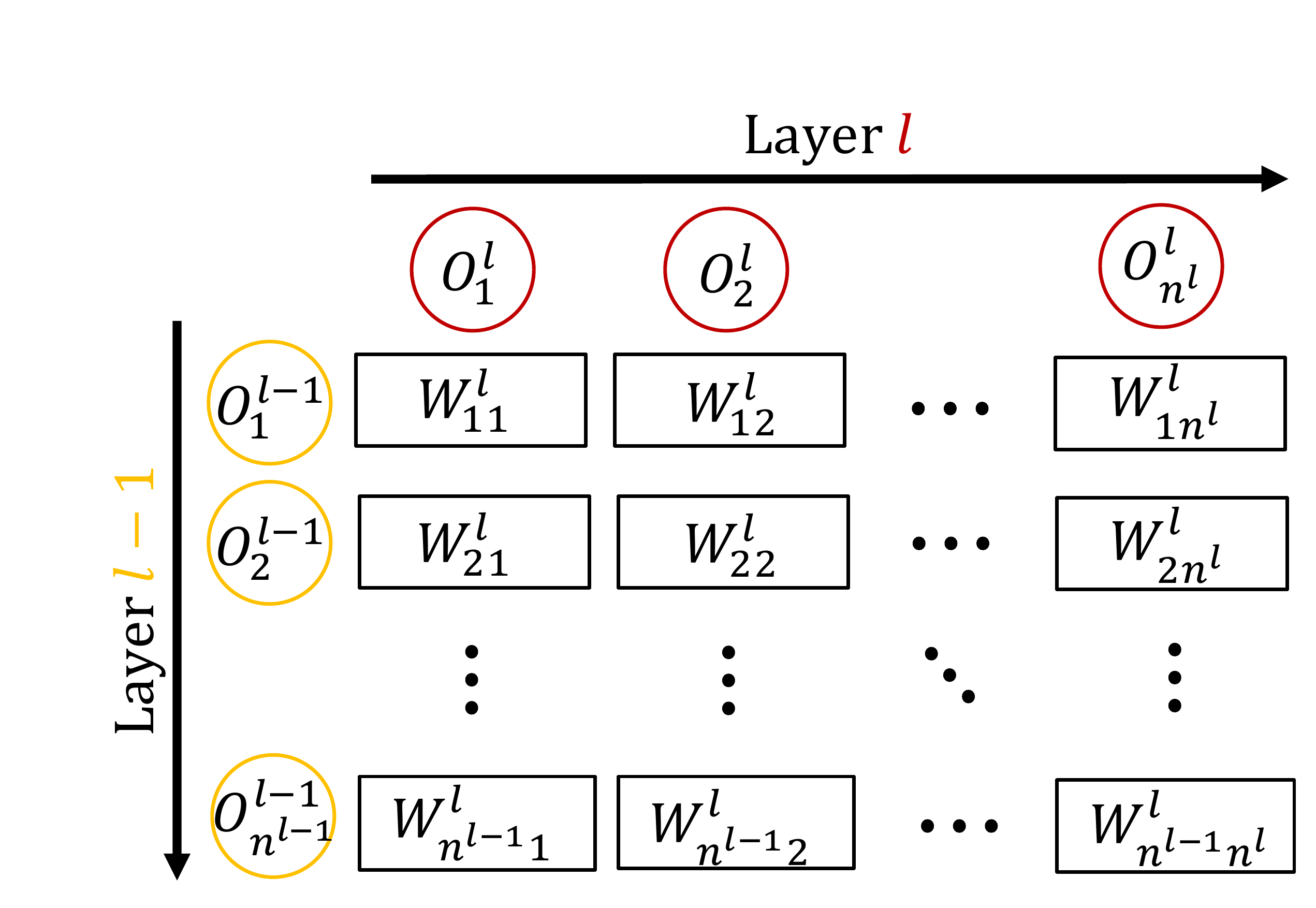}
		\caption{$\vW^\layerindex \in \R^{n^{\layerindex-1} \times n^{\layerindex}}$ from layer $\layerindex-1$ to layer $\layerindex$ }
		\label{fig:fullcolumn}
	\end{subfigure}
	~ %add desired spacing between images, e. g. ~, \quad, \qquad, \hfill etc. 
	\begin{subfigure}[b]{0.47\textwidth}
		\centering
		\includegraphics[width=\textwidth]{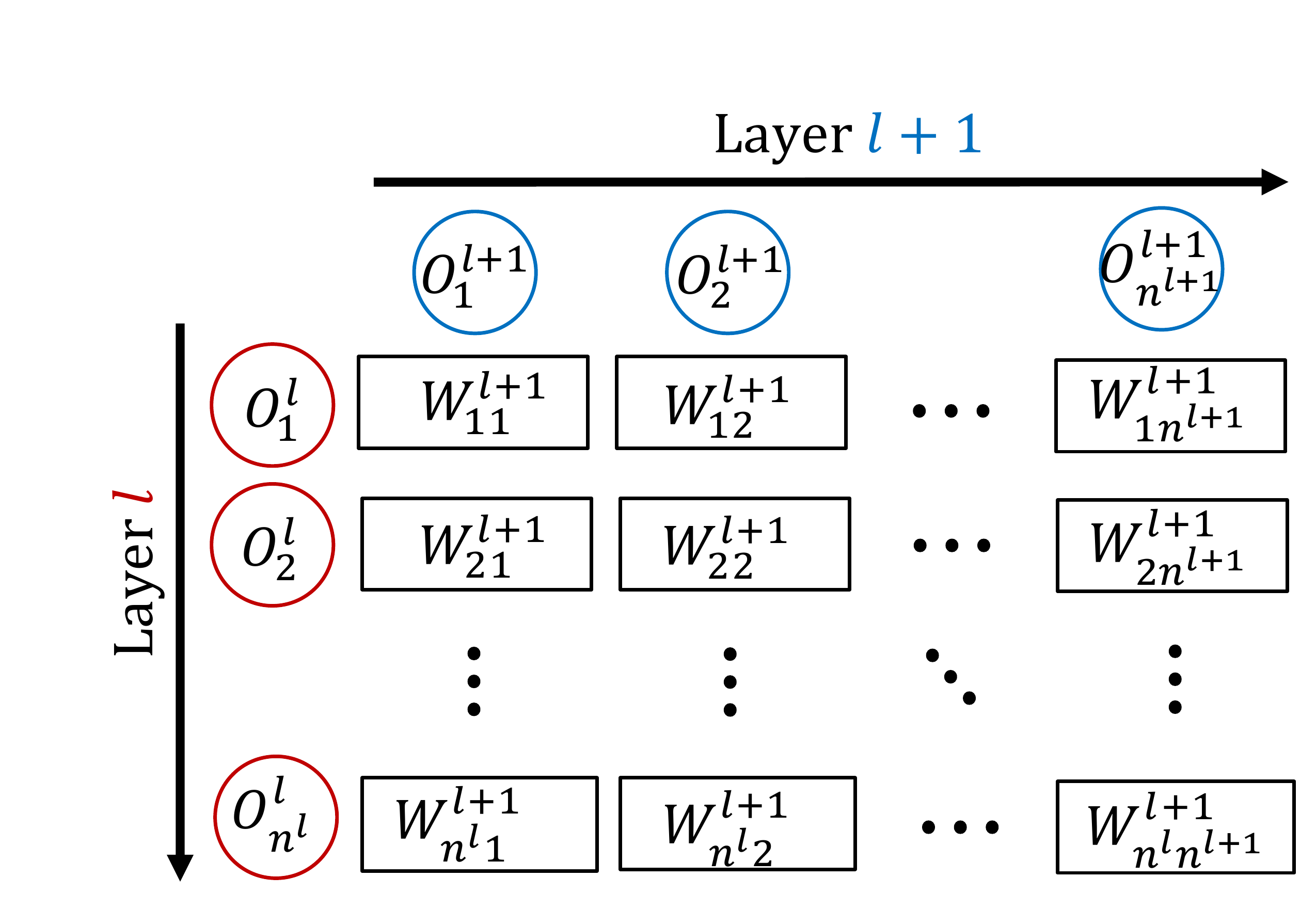}
		\caption{$\vW^{\layerindex+1} \in \R^{n^{\layerindex} \times n^{\layerindex+1 }}$ from layer $\layerindex$ to layer $\layerindex+1$ }
		\label{fig:fullrow}
	\end{subfigure}

	\begin{subfigure}[b]{0.47\textwidth}
		\centering
		\includegraphics[width=\textwidth]{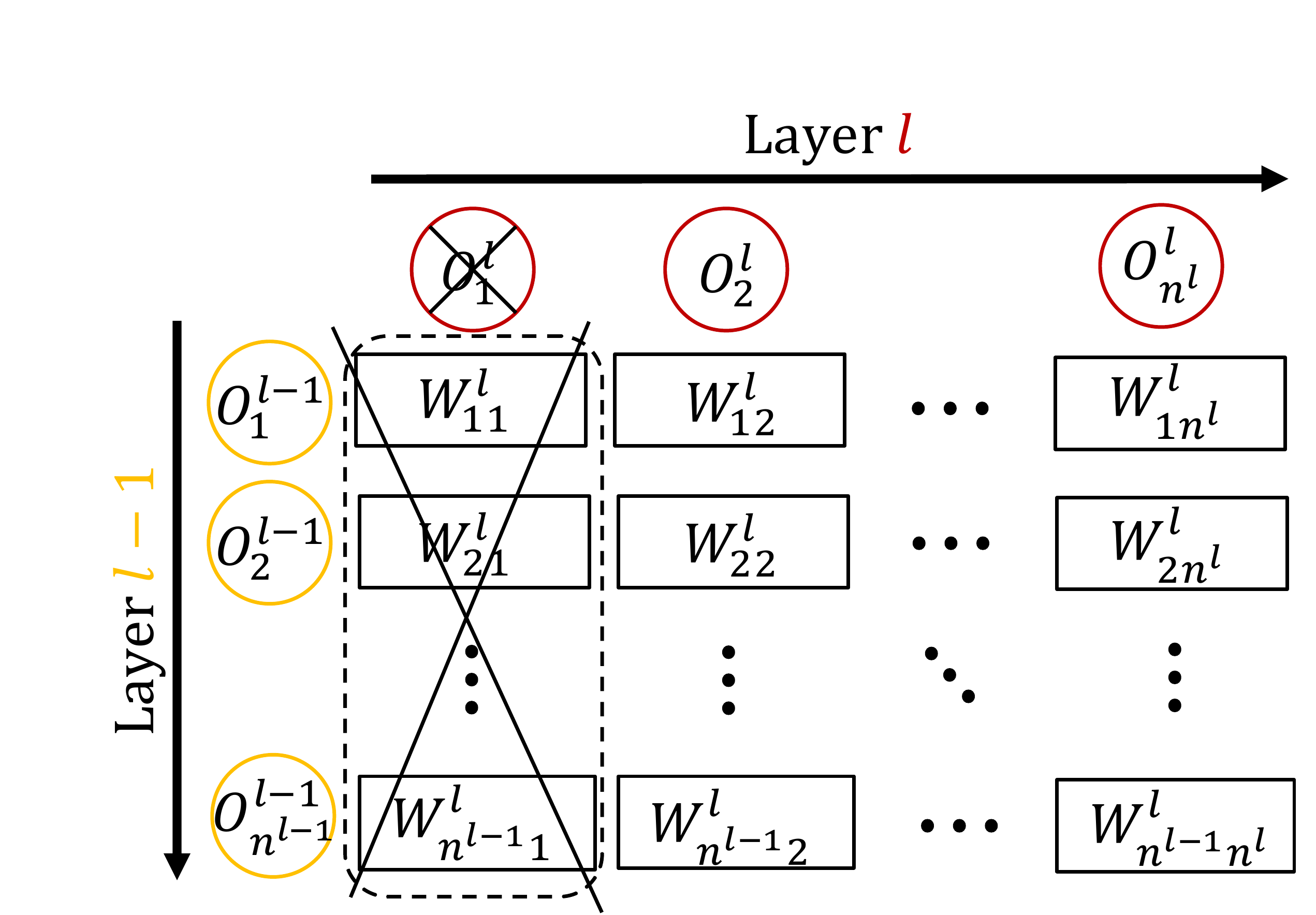}
		\caption{Removal of incoming connections to neuron $O_1^{\layerindex}$, i.e., the group of weights in the dashed box are all zeros \ \ }
		\label{fig:removecolumn}
	\end{subfigure}
	~%add desired spacing between images, e. g. ~, \quad, \qquad, \hfill etc. 
	\begin{subfigure}[b]{0.47\textwidth}
		\centering
		\includegraphics[width=\textwidth]{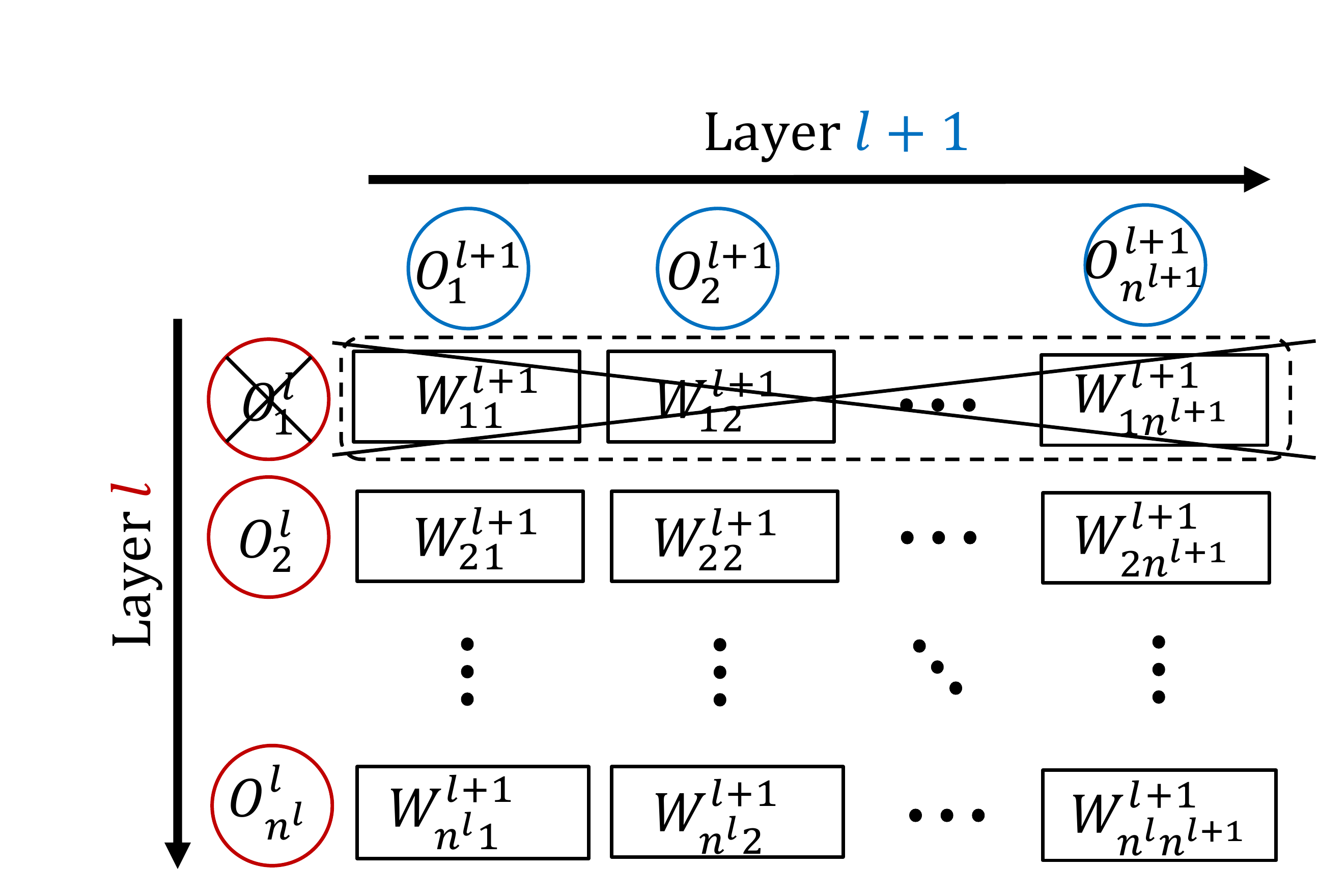}
		\caption{Removal of outgoing connections from neuron $O_1^{\layerindex}$,  i.e., the group of weights in the dashed box are all zeros }
		\label{fig:removerow}
	\end{subfigure}	
	\caption{A graphical illustration on DropNeuron strategy in Section \ref{sec:strategy}.
$O_k^{\layerindex}$ denotes the $k$-th neuron in layer ${\layerindex}$ , $W_{ij}^{\layerindex}$ denotes the weight of  connection from neuron $i$ in layer $\layerindex$ to neuron $j$ in layer $\layerindex+1$. 
		The bottom figures showed  a neuron can be removed either when all incoming connections' weights to her or her outgoing connections' weights are zeros \emph{simultaneously}.   }
	\label{fig:keyidea}
\end{figure}

\paragraph{Dropping Principle 2} 
We aim to force \texttt{li$\_$regulariser} to be small. Taking neuron $j$ in layer $\layerindex$, i.e., $O_j^{\layerindex}$ for example, \emph{all her incoming connections' weights are forced to be zeros}. It means that $O_j^{\layerindex}$  received no information from neurons in the previous layer.
Mathematically that is $\vW^\layerindex_{:,j} = \mathbf{0}$ which is valid if and only if $\Vert  \vW^\layerindex_{:,j} \Vert_2 \define \sqrt{\left(W_{1, j}^{\layerindex}\right)^2 + \ldots + \left(W_{{n^{\layerindex-1}},j}^{\layerindex}\right)^2}=0$. However, this sufficient and necessary condition is definitely not unique and can be substituted by others, e.g., dropping the root sign which becomes exactly $\ell_2$ norm $\Vert\vW^\layerindex_{:,j}\Vert_2^2$, or changing to $\ell_1$ norm $\Vert\vW^\layerindex_{:,j}\Vert_1$. 
If more than one neuron in layer $\layerindex$ are expected to be dropped, $\vW^\layerindex_{:,j}$ can be simply summed up over all $j$ as $\sum_{j=1}^{n^{\layerindex}}{ \sqrt{\sum_{i=1}^{n^{\layerindex-1}} \left(W_{ij}^{\layerindex}\right)^2}}$.
Now, it might be clear that why the square root sign can't be dropped (using $\ell_2$ norm) or replaced by $\ell_1$ norm because the independent grouping effect for \emph{all} the incoming weights of each neuron will be lost.
This idea is inspired by Group Lasso \citep{yuan2007model} to some extent, which is known to be an extension of Lasso and very well studied in statistics. Though the purpose for ours and Group Lasso is different, the regularisation norm $\Vert \cdot \Vert_2$ is the same. 
The conceptual idea of removing all the incoming weights to neuron $O_1^{\layerindex}$ from the neurons in layer $\layerindex-1$ therefore removal of herself is illustrated by comparing in Fig. \ref{fig:fullcolumn} and\ref{fig:removecolumn}. 
 
\paragraph{Dropping Principle 3} 
We aim to force \texttt{lo$\_$regulariser} to be small. 
Taking neuron $i$ in layer $\layerindex$, i.e., $O_i^{\layerindex}$ for example, \emph{all her outgoing connections' weights are forced to be zeros}. It means that $O_i^{\layerindex}$  was blocked to send information to neurons in the next layer. 
Situations of blocking exist when the outputs of several neurons in layer $\layerindex$, e.g., $O_p^{\layerindex}$ and $O_q^{\layerindex}$ are exactly the same.
Some simple examples include regression problem where the $p$-th and the $q$-th feature are exactly the same; image classification problem where pixel $p$ and pixel $q$ for all the images are exactly the same. Therefore, it may be expected that the outgoing weights from neuron $O_p^{\layerindex}$ are set to zeros and $W^{\layerindex+1}_{p,j}+W^{\layerindex+1}_{q,j}$ to be new weight from neuron $O_q^{\layerindex}$ to neuron $O_j^{\layerindex+1}$, $j = 1,\ldots, n^{\layerindex+1}$; or the other way around. 
The conceptual idea of removing all the outgoing weights from neuron $O_1^{\layerindex}$ to the neurons in layer $\layerindex+1$ therefore removal of herself is illustrated by comparing in Fig. \ref{fig:fullrow} and \ref{fig:removerow}. 

\subsubsection{New Cost Function}
Now, we can write the new cost function either for regression or classification problem
\begin{equation}
\begin{aligned}
\mathcal{L}\define \frac{1}{N} \sum_{i=1}^{N}E(\vy_i,\hat{\vy_i} ) +  \texttt{li$\_$regulariser} +\texttt{lo$\_$regulariser} 
\label{cost}
\end{aligned}
\end{equation}
or furthermore add weights regularisation term when overfitting needs to be further constrained
\begin{equation}
\begin{aligned}
\mathcal{L}\define \frac{1}{N} \sum_{i=1}^{N}E(\vy_i,\hat{\vy_i} ) +  \texttt{li$\_$regulariser} + \texttt{lo$\_$regulariser} +\texttt{l1$\_$regulariser} 
\label{cost2}
\end{aligned}
\end{equation}
The consequence of introducing new cost functions is to promote the group removal of each neuron's connections in the training process.
It should be emphasised that the regularisation parameters/hyperparameter $\lambda_{\ell_i}$ and $\lambda_{\ell_o}$ should be fine tuned carefully.

\subsection{Convex Relaxation for a ``Nearly'' Minimal Network}
The two new regularisers: \texttt{li\_regulariser} \eqref{li_regularizer} and \texttt{lo\_regulariser} \eqref{lo_regularizer} are convex functions \citep{boyd2004convex}. This 
Such convexity promises the differentiation of cost function in the training process using backpropagation in conjunction with an optimization method such as gradient descent. 

Now, we would like to replace \texttt{li$\_$regulariser} and \texttt{lo$\_$regulariser} with the following respectively
\begin{equation}
\begin{aligned}
\lambda_{\ell_o}\sum_{\layerindex=1}^{L} \sum_{i=1}^{n^{\layerindex-1}}\lVert \Vert  \vW^\layerindex_{:,j} \Vert_2\rVert_0 ,
\ \ \ \ \ \
\lambda_{\ell_i} \sum_{\layerindex=1}^{L}\sum_{j=1}^{n^{\layerindex}}\lVert { \Vert  \vW^\layerindex_{i,:} \Vert_2} \lVert_0
\label{l0_regularizer}
\end{aligned}
\end{equation}
where $\Vert \vW^\layerindex \Vert_0$ denote $\ell_0$ pseudo norm from \citep{donoho2006compressed}, i.e., the number of non-zero entries in $\vW^\layerindex$. Then we get the following cost function
\begin{equation}
\begin{aligned}
\mathcal{L}\define \frac{1}{N} \sum_{i=1}^{N}E(\vy_i,\hat{\vy_i} ) +  \lambda_{\ell_o}\sum_{\layerindex=1}^{L} \sum_{i=1}^{n^{\layerindex-1}}\lVert \Vert  \vW^\layerindex_{:,j} \Vert_2\rVert_0+
\lambda_{\ell_i} \sum_{\layerindex=1}^{L}\sum_{j=1}^{n^{\layerindex}}\lVert { \Vert  \vW^\layerindex_{i,:} \Vert_2} \lVert_0
\label{cost0}
\end{aligned}
\end{equation}
Minimisation of the cost function to the sparsest solution is generally intractable by an exhaustive search.
Therefore, we use $\Vert  \vW^\layerindex_{:,j} \Vert_2$ which is the tightest convex relaxation for $\lVert \Vert  \vW^\layerindex_{:,j} \Vert_2\rVert_0$; $\Vert  \vW^\layerindex_{i,:} \Vert_2$ which is the tightest convex relaxation for $\lVert \Vert  \vW^\layerindex_{i,:} \Vert_2\rVert_0$ as alternatives.

The solution to such convex relaxations is suboptimal to the $\ell_0$ norm solution but works well in practice and hugely facilitate the optimisation. 
Therefore, the relaxation yields a ``nearly'' sparsest solution, in other words, ``nearly'' minimal NN.
Later in Section \ref{sec:experiment}, to our surprise, the relaxed solution to the first task on sparse linear regression is almost exact compared to the true solution. We suspect that there may exist performance guarantee like restricted isometry property in compressive sensing \citep{candes2005decoding}.

\section{Experiments}
\label{sec:experiment}
Our implementation is based on the TensorFlow framework \citep{tensorflow2015-whitepaper} using GPU acceleration. The code is available on line\footnote{\code}.
At the end of training, we prune the small-weight connections: all connections with absolute weights below a threshold (typically small, e.g. $10^{-2}$) are removed from the network without reducing the test accuracy. Throughout the examples, we use the following abbreviation to indicate regularisation methods. 
$\ell_1$: $\ell_1$ regularisation, P: pruning, DO: Dropout, DN:DropNeuron, FC1: fully connected layer 1.
%\[
%\centering
%\begin{tabular}{| l | l| }
%\hline
%Abbreviation & Description\\ \hline
%$\ell_1$ & $\ell_1$ regularisation\\ \hline
%P & prune \\ \hline
%Do & Dropout \\ \hline
%DN & DropNeuron \\ \hline
%FC1 & fully connected layer 1 \\ \hline
%\end{tabular}
%\]

\subsection{Sparse Regression}

We started with a simple sparse linear regression problem which is a classic problem in compressive sensing or sparse signal recovery.
The inputs and outputs were synthetically generated as follows.
First, a random feature matrix $\Phi \in \R ^{m\times n}$, often overcomplete, was created whose columns are each drawn uniformly from the surface of the unit sphere in $\R^n$. Next, sparse coefficient vectors $x_0 \in \R ^{n}$ are randomly generated with $d$ nonzero entries. Nonzero magnitudes $\bar{x}_0$ are drawn i.i.d. from an experiment-dependent distribution. Signals are then computed as $y = \Phi x_0 \in \R ^{m}$, and then contaminated by adding noise $\xi \in \R ^{m}$ with certain distribution. i.e., $y = \Phi x_0 + \xi$.  In compressive sensing or sparse signal recovery setting, several algorithms will be presented with $y$ and $\Phi$  and attempts to estimate $x_0$. Such training can be formulated by a neural network where an extreme case will be there is only one hidden layer and there is only one neuron on thin layer. Minimisation of a cost function with mean square error as loss and $\ell_1$ as regulariser over the weight will typically yield the exact solution if $\Phi$ satisfy conditions like restricted isometry property \citep{candes2005decoding}.

Rather than using a single hidden layer and single neuron for training, we specified a multi-layer structure and there are more than one neurons in each layer.
To be simple, the activation function is assumed to be linear.
Therefore, the training of $x_0$ is not the main concern under the deep neural network framework but the prediction error for the test set is more interesting. In our experiment, 
the number of example in training set and test set are the same.
We used the standard normalised mean square error (NMSE)  metric , i.e. $\text{NMSE} = \frac{\sum_{t=1}^N(y_t-\hat{y}_t)^2}{\sum_{t=1}^{N}y_t^2}$, to evaluate the prediction accuracies of the models.

It seems that deep neural architecture with multiple layers and many neurons is overly used for this simple example. It should be naturally expected that the prediction error is as small as possible especially after adding regularisation technique such as Dropout. However, the results seems to be counter-intuitive while our method yield impressive performance. 

First of all, we set the number of features $n$ to be $20$ and there are $2$ nonzero elements in $x_0$.
Only one hidden layer is specified, with $5$ neurons in this layer. Therefore, $\vW^1 \in \R^{20\times 5}$ and the output layer $\vW^2 \in \R^{5}$ .%, with the similar structure in Fig. \ref{fig:weight}. 
After each layer, we applied Dropout with a keeping probability of $50\%$. 
The number of example was set to be $1000$ (half for training and half for testing) which is much greater than the number of unknown weight ($20\times 5 + 5 = 105$). 
The setup of experiment was as follows:  
optimizer: \texttt{AdamOptimizer};
number of epochs: 100;
learning rate: 0.001;
batch size: 1;
dropout keep probability : 50\%.

%\[
%\centering
%\begin{tabular}{| l | l| }
%\hline
%Optimizer & \texttt{AdamOptimizer} \\ \hline
%Number of epochs & 100\\ \hline
%Learning rate & 0.001 \\ \hline
%Batch size& 1 \\ \hline
%Dropout keep probability & 50\% \\ \hline
%\end{tabular}
%\]
In all cases, we ran $1000$ independent trials to generate different feature matrix and output.
As an illustration, we show the training result in one trial where the prediction NMSE using Dropout is the lowest among all the trials. 
In this trail,  the spare vector $x_0 = [ 0, 0, 3.87308349,  0, 0, 0, 0,  0, 0,  -8.23781791,  0, 0, 0, 0, 0,  0,  0,  0,  0,  0]$, where the $3^{rd}$ and $10^{th}$ entries are nonzeros. 
The estimated weights using Dropout are shown in Appendix \ref{app:example-1}, both $\vW^1$ in \eqref{app:w1_dropout} and $\vW^2$ in \eqref{app:w2_dropout} are not sparse and implying a fully connected architecture. 
The test NMSE is around $0.54$.

Using the same data, the training result using DropNeuron can be found in Appendix \ref{app:example-1}, both $\vW^1$ in \eqref{W1_DropNeuron} and $\vW^2$ in \eqref{W2_DropNeuron} are very sparse. In $\vW^1$, only two non zeros weights are found, they are 
$W^1_{3, 2}=-0.6687693$ and $W^1_{10,2} = 1.42591035$; and in $\vW^2$, there is only one nonzero entry $\vW^2_2 = -5.74600601$. The test NMSE is surprisingly low at around $0.00036$.

%{\small
%\begin{equation}
%\begin{aligned}
%\vW^1_{DropNeuron} = 
%\begin{bmatrix}
%0 &          0 &         0 &         0    &      0        \\
%0 &         0   &       0  &        0   &       0       \\
%0 &        -0.6687693  & 0      &    0     &     0        \\
%0 &         0   &       0  &        0    &      0        \\
%0 &         0  &        0   &       0    &      0   \\     
%0  &        0   &       0  &        0   &       0        \\
%0  &        0   &       0  &        0   &       0        \\
%0  &        0   &       0   &       0   &       0        \\
%0  &        0  &        0   &       0   &       0        \\
%0  &        1.42591035 & 0 &         0  &        0   \\     
%0  &        0  &        0  &        0    &      0        \\
%0  &        0 &         0  &        0    &      0        \\
%0  &        0 &         0  &        0    &      0        \\
%0   &       0 &         0  &        0    &      0        \\
%0   &       0 &         0  &        0    &      0        \\
%0  &        0 &         0  &        0    &      0        \\
%0 &         0  &        0  &        0    &      0        \\
%0  &        0  &        0  &        0    &      0        \\
%0  &        0  &        0  &        0    &      0        \\
%0  &        0  &        0  &        0   &       0        
%\end{bmatrix}
%\label{W1_DropNeuron }
%\end{aligned}
%\end{equation}
%}
%and
%\begin{equation}
%\begin{aligned}
%\vW^2_{DropNeuron} = 
%\begin{bmatrix}
%0        \\
%-5.74600601\\
% 0        \\
%0      \\
%0        
%\end{bmatrix}
%\label{W2_DropNeuron }
%\end{aligned}
%\end{equation}

It is a fact that the only two nonzero entries of $\vW^1$ both appear in the second column of $\vW^1$. This means that only the second neuron in the hidden layer is necessary to be kept while dropping all the other neurons. Similarly, the second neuron in the output layer is necessary to exist. Meanwhile, we notice that $W^1_{3, 2} \times W^2_2 = (-0.6687693) \times (-5.74600601) = 3.8427524171$ and $W^1_{10, 2} \times W^2_2 =1.42591035 \times (-5.74600601)= -8.19328944082$, which are very close to the nonzero entry in  $x_0 = [ 0, 0, 3.87308349,  0, 0, 0, 0,  0, 0,  -8.23781791,  0, 0, 0, 0, 0,  0,  0,  0,  0,  0]$. If we investigate the structure of \eqref{W1_DropNeuron} and \eqref{W2_DropNeuron} again, and considering the effect of linear activation function, the estimated network architecture by dropping unnecessary neurons almost reveal the true additive structure of the third and tenth feature. A conceptual illustration for the strategy of dropping neurons for the regression problem can be found in Fig. \ref{fig:droplinearregression}.

\begin{table}[t]
\centering
%\vspace{-30pt}
\caption{Summary of statistics for Sparse Regression (best NMSE using DO) }
\resizebox{\textwidth}{!}
{%
\begin{tabular}{l|lllll}
\hline
Regularisation & $\vW^\text{FC1}\%$ & $\vW^\text{FC2}\%$  & $\vW^{\text{total}} \%$  & NMSE  & NMSE (no prune)  \\ \hline
$\ell_1$+DO+P & $58\%$ & $100\%$ & $60\%$   & 0.54& 0.54\\
$\ell_1$+DN+P & $16.00\%$ & $44.47\%$ & $54.11\%$  & 0.00036 &0.00036\\
\hline
Regularisation & $\vO^\text{input}\%$ & $\vO^\text{FC1}\%$ & $\vO^\text{output}\%$ & $\vO^{\text{total}} \%$ & Compression Rate  \\ \hline
$\ell_1$+DO+P & $\frac{20}{20} = 100\%$& $\frac{5}{5} = 100\%$   & $\frac{1}{1} = 100\%$& $\frac{26}{26} = 100\%$ & 1.67  \\
$\ell_1$+DN+P & $\frac{2}{20} = 10\%$& $\frac{1}{5} = 20\%$   & $\frac{1}{1} = 100\%$& $\frac{4}{26} = 15.38\%$ & 35  \\
\hline
\end{tabular}
}
\label{table:regression}
\end{table}

\subsection{Deep Autoencoder}
\begin{table}[t]
\centering
%\vspace{-30pt}
\caption{Summary of statistics for Autoencoder (average over 10 initialisations).} 
%Abbreviation: $\ell_1$: $\ell_1$ regularisation, P: pruning, DO: Dropout, DN:DropNeuron.}
\resizebox{\textwidth}{!}
{%
\begin{tabular}{l|lllllllc}
\hline
Regularisation & $\vW^\text{FC1}\%$ & $\vW^\text{FC2}\%$ & $\vW^\text{FC3}\%$ & $\vW^\text{FC4}\%$ & $\vW^{\text{total}} \%$  & NMSE  & NMSE (no prune)  \\ \hline
DO+P & $99.57\%$ & $99.39\%$ & $99.45\%$  & $99.60\%$& $99.58\%$ & 0.031 & 0.031\\
$\ell_1$+DO+P & $15.18\%$ & $46.29\%$ & $52.53\%$  & $17.54\%$& $18.86\%$ & 0.011& 0.011\\
$\ell_1$+DN+P & $16.00\%$ & $44.47\%$ & $54.11\%$  & $18.14\%$& $19.50\%$ & 0.012 &0.012\\
\hline
Regularisation & $\vO^\text{input}\%$ & $\vO^\text{FC1}\%$ & $\vO^\text{FC2}\%$ & $\vO^\text{FC3}\%$ & $\vO^\text{output}\%$ & $\vO^{\text{total}} \%$ & Compression Rate  \\ \hline
DO+P & $\frac{784}{784} = 100\%$& $\frac{128}{128} = 100\%$ & $\frac{64}{64} = 100\%$ & $\frac{128}{128} = 100\%$  & $\frac{784}{784} = 100\%$& $\frac{1888}{1888} = 100\%$ & 1.0  \\
$\ell_1$+DO+P & $\frac{459}{784} = 58.55\%$& $\frac{127}{128} = 99.22\%$ & $\frac{62}{64} = 96.88\%$ & $\frac{121}{128} = 94.53\%$  & $\frac{784}{784} = 100\%$& $\frac{1553}{1888} = 82.26\%$ & 5.3  \\
$\ell_1$+DN+P & $\frac{420}{784} = 53.57\%$& $\frac{127}{128} = 99.22\%$ & $\frac{61}{64} = 95.31\%$ & $\frac{121}{128} = 94.53\%$  & $\frac{629}{784} = 80.23\%$& $\frac{1358}{1888} = 71.93\%$ & 5.1\\
\hline
\end{tabular}
}
\label{table:ae}
\end{table}

We considered the image dataset of MNIST \citep{lecun1998mnist}. The number of training examples and test examples are $60000$ and $10000$ respectively, the image sizes are $28\times 28$ digit images and $10$ classes.
We used a $784\rightarrow 128\rightarrow64 \rightarrow 128 \rightarrow 784$ autoencoder and all units were logistic with mean square error as loss. Let's train the autoencoder for 50 epochs. After 50 epochs, we try to visualise the reconstructed inputs and the encoded representation without using different combination of regularisations as showed in Fig. \ref{fig:aedigit}. 

In Fig. \ref{fig:W1}, we illustrated the sparsity pattern of the estimated weight matrix by setting all the nonzeros weights to one instead of the true value. It is shown in Fig. \ref{fig:W1i} that the top and bottom rows are all zeros. It means that the corresponding features/pixels of the image have no effect to the subsequent layer. During the training and testing process, the patch in Fig. \ref{fig:aedigit1} is typically vectorised where the left and right blue pixels are inputs targeting the top and bottom area of the weight matrix. Apparently, these blue pixels are picture background without any useful information. Furthermore, a conceptual illustration on DropNeuron strategy can be found in Fig. \ref{fig:ae}.

A summary of training and testing statistics can be found in Table \ref{table:ae}. 
It is ambitiously to expect DropNeuron can yield low NMSE, low total sparsity level/high compression rate, few neurons simultaneously.
It turns out that total sparsity level/compression rate/NMSE is slightly higher/lower/higher using DropNeuron than Dropout together with $\ell_1$ regularisation and pruning. However the number of neurons dropped using DropNeuron is much higher. This may be due to the use of mean square error metric as loss function to be minimised. The training process is trying to recover the input image. We can't guarantee lower NMSE involves no fitting to noise. And more importantly, we need to consider the unsupervised nature of this task: feature representation. Dropping more neurons could give insight to the extracted features. In the next example on convolutional NN for classification, DropNeuron outperforms in all aspects for the supervised learning task.

\subsection{Convolutional NN}

It is well known that the fully connected layer is ``parameter intensive'' (more than $90\%$ of the model size). This typically raised a problem to store too many parameters to store on one single machine but across multiple ones. A consequence is the communication among machines which is an inhibitor to the computation efficiency. 
In this example, consider the LeNet-5 with two convolutional layers and two fully connected layers for classification of MNIST dataset. 

It should be noted that we are not competing with the state-of-art accuracy due to the various network structures with exhaustive tuning of hyperparameters, such as batch size, initial weights, learning rate, etc. We would like to demonstrate that models trained with DropNeuron regularisation can achieve comparable (better) accuracy with ones trained with other regularisations such as Dropout, $\ell_1$ regularisation, etc, while fixing the other conditions. In Fig. \ref{fig:1FC} and Fig. \ref{fig:2FC}, we illustrated the actual training weights and their sparsity patterns of the fully connected layers under combinations of various regularisations. A summary of training and testing statistics can be found in Table \ref{table:lenet5}. It should be noted that our approach is unable to drop the neurons of filters ($25\times32+25\times 32 \times 64 = 52000$ neurons in LeNet-5) in the convolutional layers. Even so, the compression rate is above $60 \%$.

\begin{table}[h]
\centering
%\vspace{-30pt}
\caption{Summary of statistics for the fully connect layer of LeNet5 (average over 10 initialisations) }
\resizebox{\textwidth}{!}
{%
\begin{tabular}{l|lllll}
\hline
Regularisation & $\vW^\text{FC1}\%$ & $\vW^\text{FC2}\%$  & $\vW^{\text{total}} \%$  & Accuracy  & Accuracy (no prune)  \\ \hline
DO+P & $55.15\%$ & $62.81\%$ & $55.17\%$   & $99.07\%$&$99.12\%$\\
$\ell_1$+DO+P & $5.42\%$ & $51.66\%$ & $5.57\%$   & $99.01\%$& $98.96\%$\\
$\ell_1$+DN+P & $1.44\%$ & $16.82\%$ & $1.49\%$  & $99.07\%$&$99.14\%$\\
\hline
Regularisation & $\vO^\text{FC1}\%$ & $\vO^\text{FC2}\%$ & $\vO^\text{output}\%$ & $\vO^{\text{total}} \%$ & Compression Rate  \\ \hline
DO+P & $\frac{3136}{3136} = 100\%$& $\frac{504}{512} = 98.44\%$   & $\frac{10}{10} = 100\%$& $\frac{3650}{3658} = 99.78\%$ & 1.81 \\
$\ell_1$+DO+P & $\frac{1039}{3136} = 33.13\%$& $\frac{320}{512} = 62.5\%$   & $\frac{10}{10} = 100\%$& $\frac{1369}{3658} =37.42\%$ & 17.95 \tablefootnote{The result is consistent with \citep{han2015learning, han2015deep_compression} just using pruning}  \\
$\ell_1$+DN+P & $\frac{907}{3136} = 28.92\%$& $\frac{110}{512} =21.48\%$   & $\frac{10}{10} = 100\%$& $\frac{1027}{3658} = 28.08\%$ & 67.04 \\
%\tablefootnote{$1.5\times$ better than the best result in \citep{ han2015deep_compression}}  \\
\hline
\end{tabular}
}
\label{table:lenet5}
\end{table}

\section{Conclusions and Future Research}

We presented a novel approach of optimising a deep neural network through regularisation of network architecture. We proposed regularisers which support a simple mechanism of dropping neurons during a network training process. The method supports the construction of a simpler deep neural networks with compatible performance with its simplified version.   We evaluate the proposed method with few examples including sparse linear regression, deep autoencoding and convolutional net. The valuations demonstrate excellent performance.

This research is in its early stage. First, we have noticed that for specific deep NN structures such as Convolutional NN, Recurrent NN, Restricted Boltzmann Machine, etc, the regularisers need to be adjusted respectively. Second, we also notice that Dropout training in deep NN as approximate Bayesian inference in deep Gaussian processes which offer a mathematically grounded framework to reason about model uncertainty \citep{Gal2015DropoutB}.  Both \texttt{lo$\_$regulariser} and \texttt{li$\_$regulariser} may be potentially explained from Bayesian perspective by introducing specific kernel functions \citep{rasmussen2006gaussian}.

\newpage
\appendix
\begin{center}
{\Large \textbf{Supplementary Material}}
\end{center}
\numberwithin{equation}{section}
\setcounter{figure}{0}
\renewcommand\thefigure{S.\arabic{figure}}

\section{Example: Sparse Linear Regression}
\label{app:example-1}

\begin{figure}[h]
	\centering
	\begin{subfigure}[b]{0.21\textwidth}
		\centering
		\includegraphics[width=\textwidth]{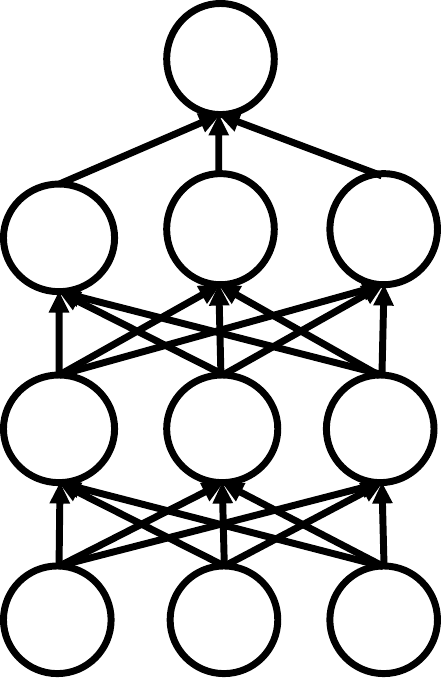}
		\caption{Full size model}
	\end{subfigure}
	~ %add desired spacing between images, e. g. ~, \quad, \qquad, \hfill etc. 
	\begin{subfigure}[b]{0.21\textwidth}
		\centering
		\includegraphics[width=\textwidth]{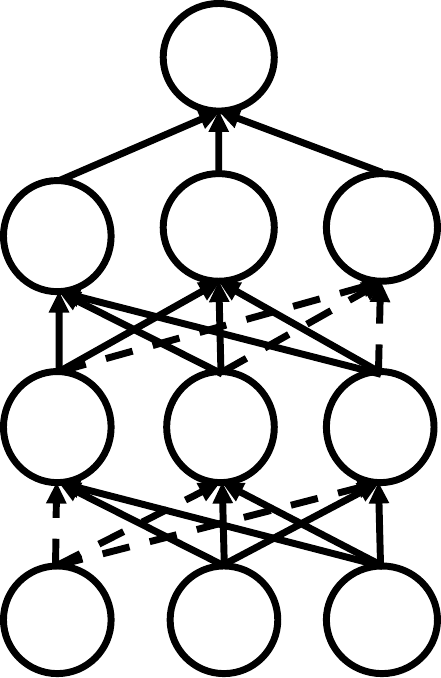}
		\caption{Drop Neurons}
	\end{subfigure}
	~
	\begin{subfigure}[b]{0.21\textwidth}
		\centering
		\includegraphics[width=\textwidth]{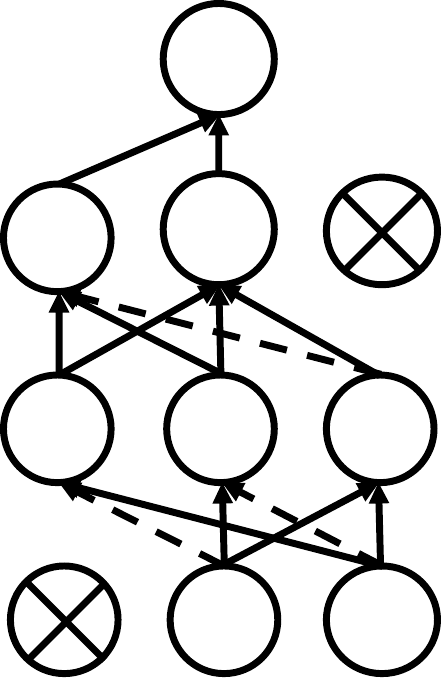}
		\caption{Drop Connections}
	\end{subfigure}
	~%add desired spacing between images, e. g. ~, \quad, \qquad, \hfill etc. 
	\begin{subfigure}[b]{0.21\textwidth}
		\centering
		\includegraphics[width=\textwidth]{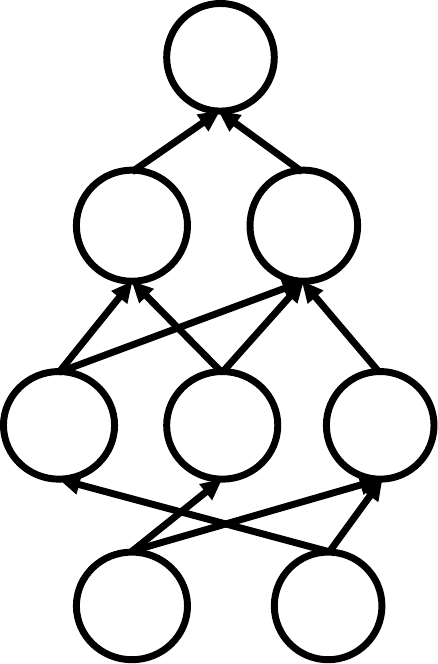}
		\caption{Small size NN}
	\end{subfigure}	
	\caption{A graphical illustration of DropNeuron strategy in regression problem}
	\label{fig:droplinearregression}
\end{figure}

\begin{equation}
\begin{aligned}
\vW^1_{Dropout}= 
\begin{bmatrix}
 0     &     0    &      0  &       -0.01453323 & 0.06075698\\
-0.05635324 & 0   &      -0.02587643 & 0.02911515 &-0.01041718\\
 0.02563882 & 0.05031364 & 0.03376988&  0.01993434 &-0.03179494\\
 0.06203449&  0.02862295& -0.06700613&  0.02385385& -0.02911432\\
-0.48254526 &-0.35333461&  0.31129083&  0.3545627 & -0.31979144\\
 0.03181251& -0.07274029 & 0.05249952&  0.04767575& -0.02953613\\
 0.04364435 & 0  &       -0.03578129& -0.03502097 & 0.09711245\\
-0.04102893& -0.06275055 & 0   &      -0.06409876 &-0.05218389\\
 0.0200471 &  0.06717232 & 0    &      0.02837713 & 0.03758603\\
-0.03055474 & 0.0289463 &  0.06301561 & 0.03308195&  0.01662179\\
-0.02746344 & 0.07223324&  0.04476647 & 0.01322776&  0.04655014\\
-0.01112585 & 0   &      -0.037157 &  -0.03381626 & 0.02151454\\
 0.04563131 &-0.03387317& -0.04606552&  0     &     0.01086553\\
 0.03301461 &-0.02328412 & 0.0114607&  -0.01552058&  0        \\
-2.000736 &   -1.94960344&  2.02926755 & 1.93757319& -1.96851373\\
-0.05102381 & 0.02301042& -0.07785907&  0.01081117&  0.0626013 \\
 0.02743321 & 0.03834696&  0.06928469&  0   &       0        \\
-0.04644512 & 0.  &       -0.01497171 & 0.02810199 & 0        \\
 0.0628076 &  0.04429785&  0.01758143&  0.01070064& -0.02718436\\
 0    &     0   &      -0.0419367 &  0.06928124& -0.05641071
\end{bmatrix}
\label{app:w1_dropout}
\end{aligned}
\end{equation}

\begin{equation}
\begin{aligned}
\vW^2_{Dropout} = 
\begin{bmatrix}
-0.10229997\\
-0.11288397\\
 0.11892998\\
 0.12453081\\
-0.11404949
\end{bmatrix}
\label{app:w2_dropout}
\end{aligned}
\end{equation}

\begin{equation}
\begin{aligned}
\vW^1_{DropNeuron} = 
\begin{bmatrix}
0 &          0 &         0 &         0    &      0        \\
0 &         0   &       0  &        0   &       0       \\
0 &        -0.6687693  & 0      &    0     &     0        \\
0 &         0   &       0  &        0    &      0        \\
0 &         0  &        0   &       0    &      0   \\     
0  &        0   &       0  &        0   &       0        \\
0  &        0   &       0  &        0   &       0        \\
0  &        0   &       0   &       0   &       0        \\
0  &        0  &        0   &       0   &       0        \\
0  &        1.42591035 & 0 &         0  &        0   \\     
0  &        0  &        0  &        0    &      0        \\
0  &        0 &         0  &        0    &      0        \\
0  &        0 &         0  &        0    &      0        \\
0   &       0 &         0  &        0    &      0        \\
0   &       0 &         0  &        0    &      0        \\
0  &        0 &         0  &        0    &      0        \\
0 &         0  &        0  &        0    &      0        \\
0  &        0  &        0  &        0    &      0        \\
0  &        0  &        0  &        0    &      0        \\
0  &        0  &        0  &        0   &       0        
\end{bmatrix}
\label{W1_DropNeuron}
\end{aligned}
\end{equation}

\begin{equation}
\begin{aligned}
\vW^2_{DropNeuron} = 
\begin{bmatrix}
0        \\
-5.74600601\\
 0        \\
0      \\
0        
\end{bmatrix}
\label{W2_DropNeuron}
\end{aligned}
\end{equation}

\section{Example: Deep Autoencoder}
\label{app:example-2}
\begin{figure} [h]
	\centering
	\begin{subfigure}[b]{0.9\textwidth}
		\centering
		\includegraphics[width=\textwidth]{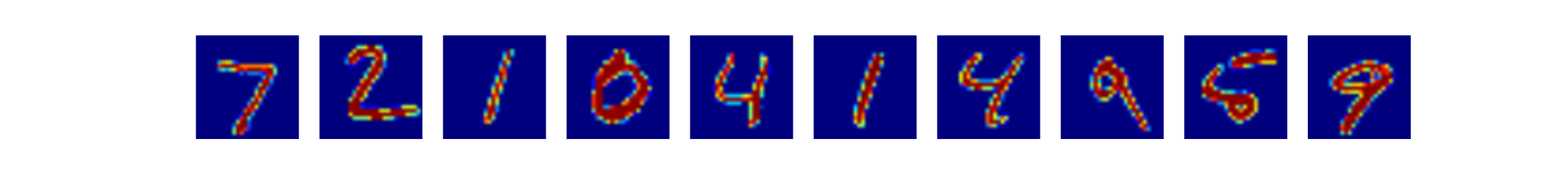}
		\caption{Original digits}
        \label{fig:aedigit1}
	\end{subfigure}
	
	\begin{subfigure}[b]{0.9\textwidth}
		\centering
		\includegraphics[width=\textwidth]{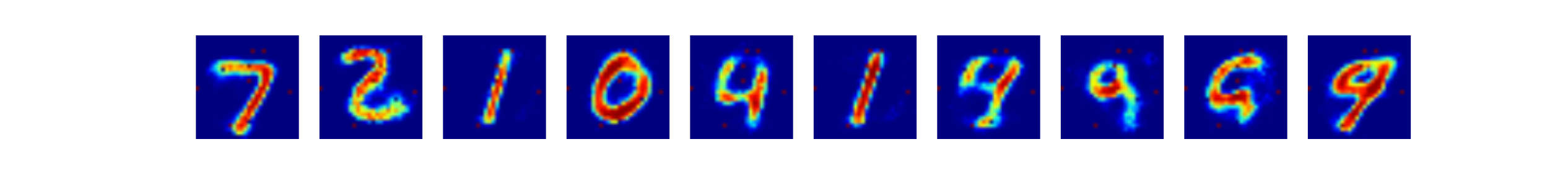}
		\caption{Reconstructed digit (DO+P), NMSE = 0.0311838}
		\label{fig:aedigit2}
	\end{subfigure}
	
	\begin{subfigure}[b]{0.9\textwidth}
		\centering
		\includegraphics[width=\textwidth]{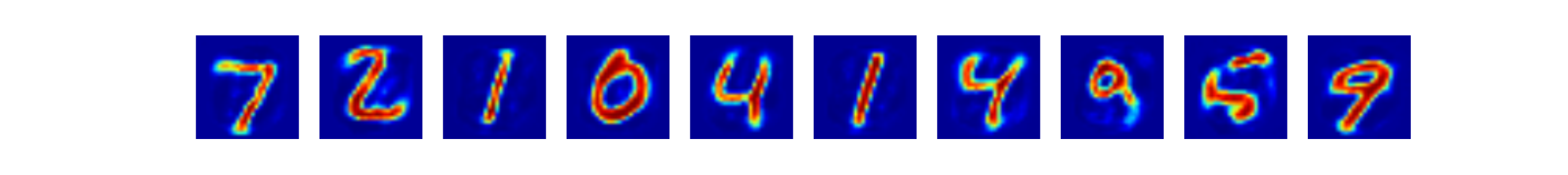}
		\caption{Reconstructed digit ($\ell_1$+DO+P), NMSE = 0.0115109}
		\label{fig:aedigit3}
	\end{subfigure}

	\begin{subfigure}[b]{0.9\textwidth}
		\centering
		\includegraphics[width=\textwidth]{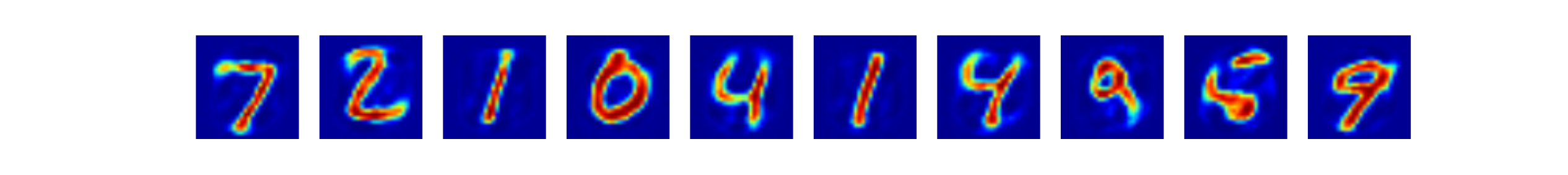}
		\caption{Reconstructed digit ($\ell_1$+DN+P), NMSE = 0.0121184} 
		\label{fig:aedigit4}
	\end{subfigure}
	\caption{The top row is the original digits, the second to the fourth rows are the reconstructed digit using deep autoencoder ($784\rightarrow 128\rightarrow64 \rightarrow 128 \rightarrow 784$) with different combination of regularisations.
	}\label{fig:aedigit}
\end{figure}

\begin{figure}[h]
	\centering
	\begin{subfigure}[b]{0.2\textwidth}
		\centering
		\includegraphics[width=\textwidth]{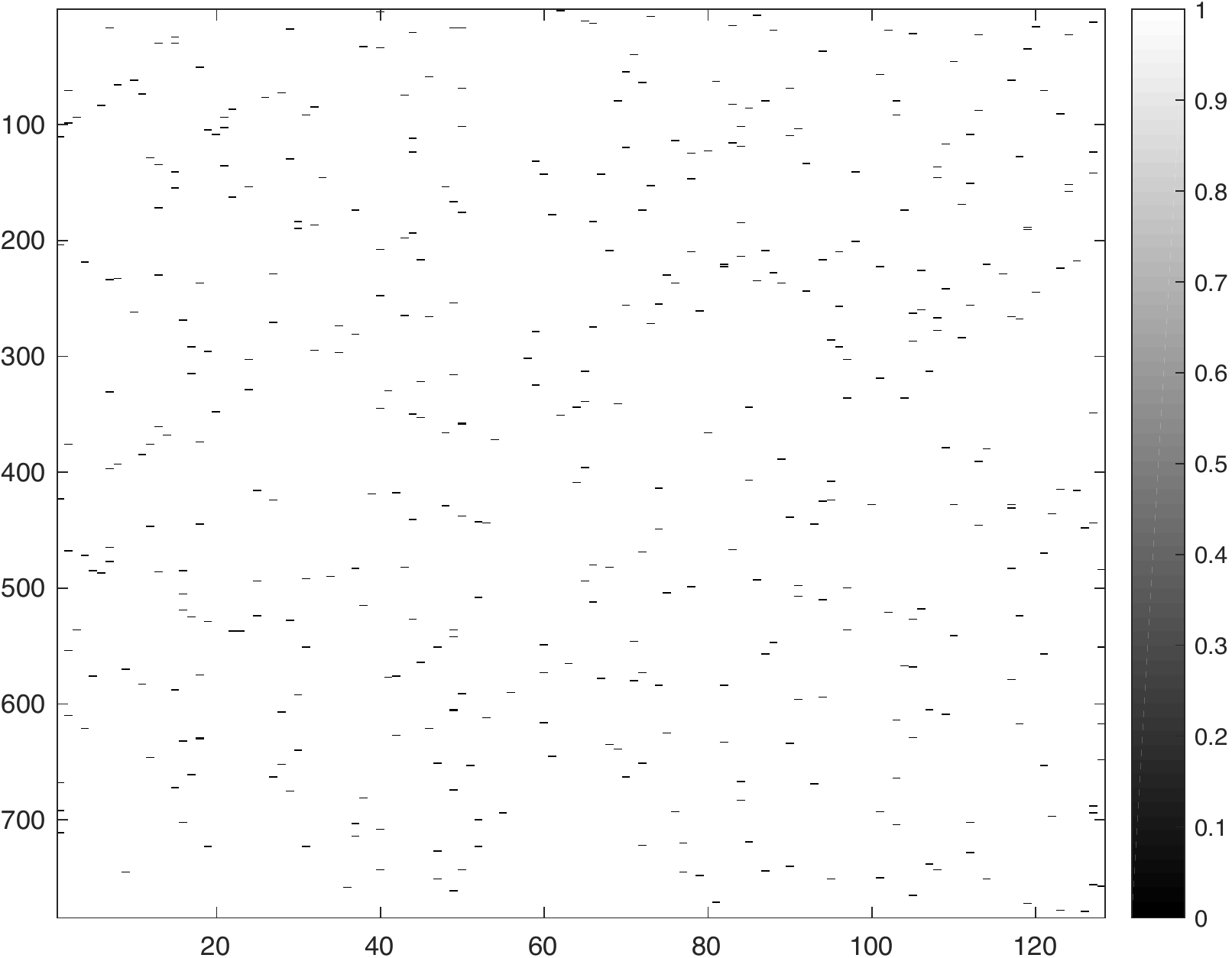}
		\caption{$\vW^\text{FC1}$ (DO+P)}
		\label{fig:W1a} 
	\end{subfigure}
~	
	\begin{subfigure}[b]{0.2\textwidth}
		\centering
		\includegraphics[width=\textwidth]{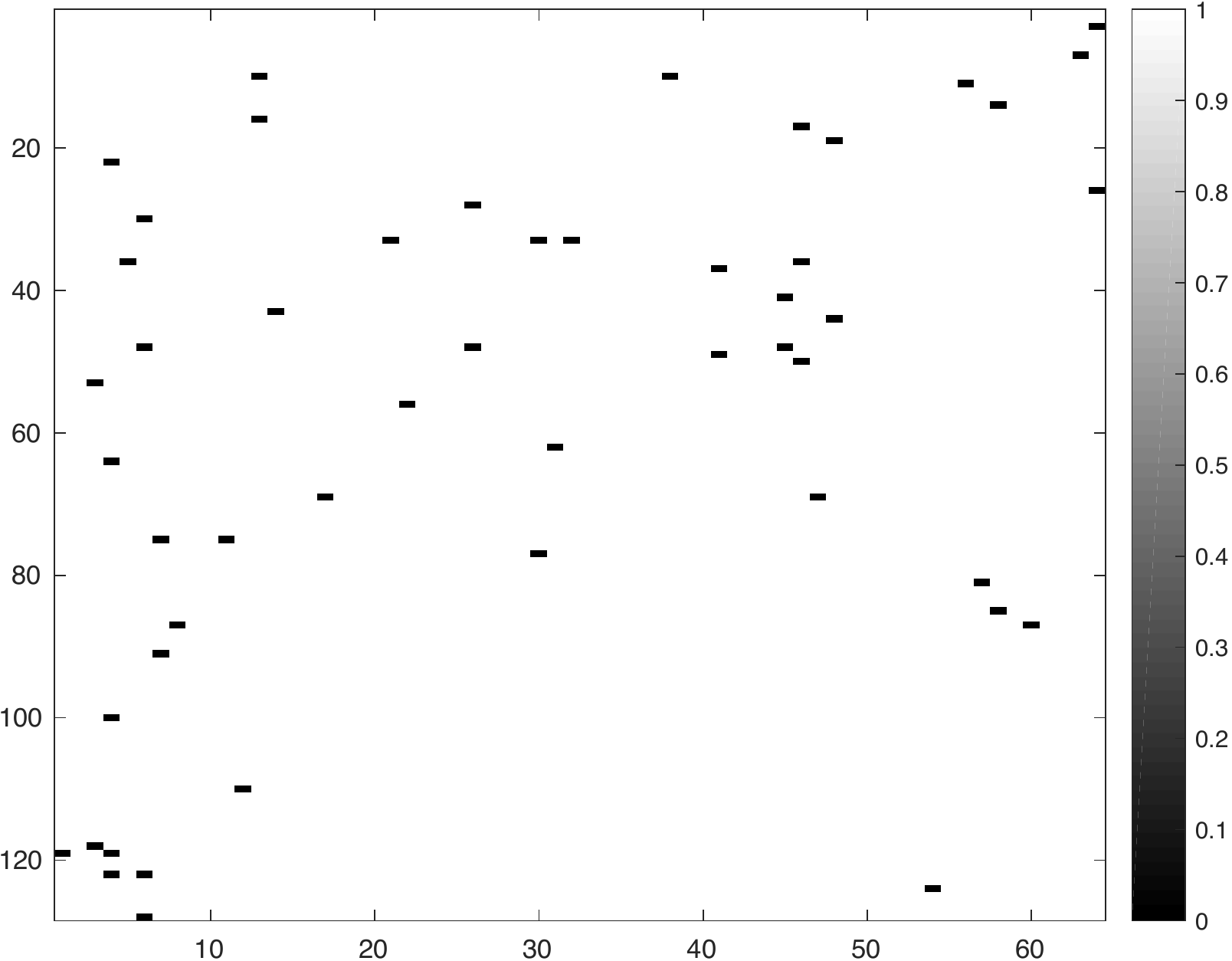}
		\caption{$\vW^\text{FC2}$ (DO+P)}
		\label{fig:W1b} 
	\end{subfigure}
~	
	\begin{subfigure}[b]{0.2\textwidth}
		\centering
		\includegraphics[width=\textwidth]{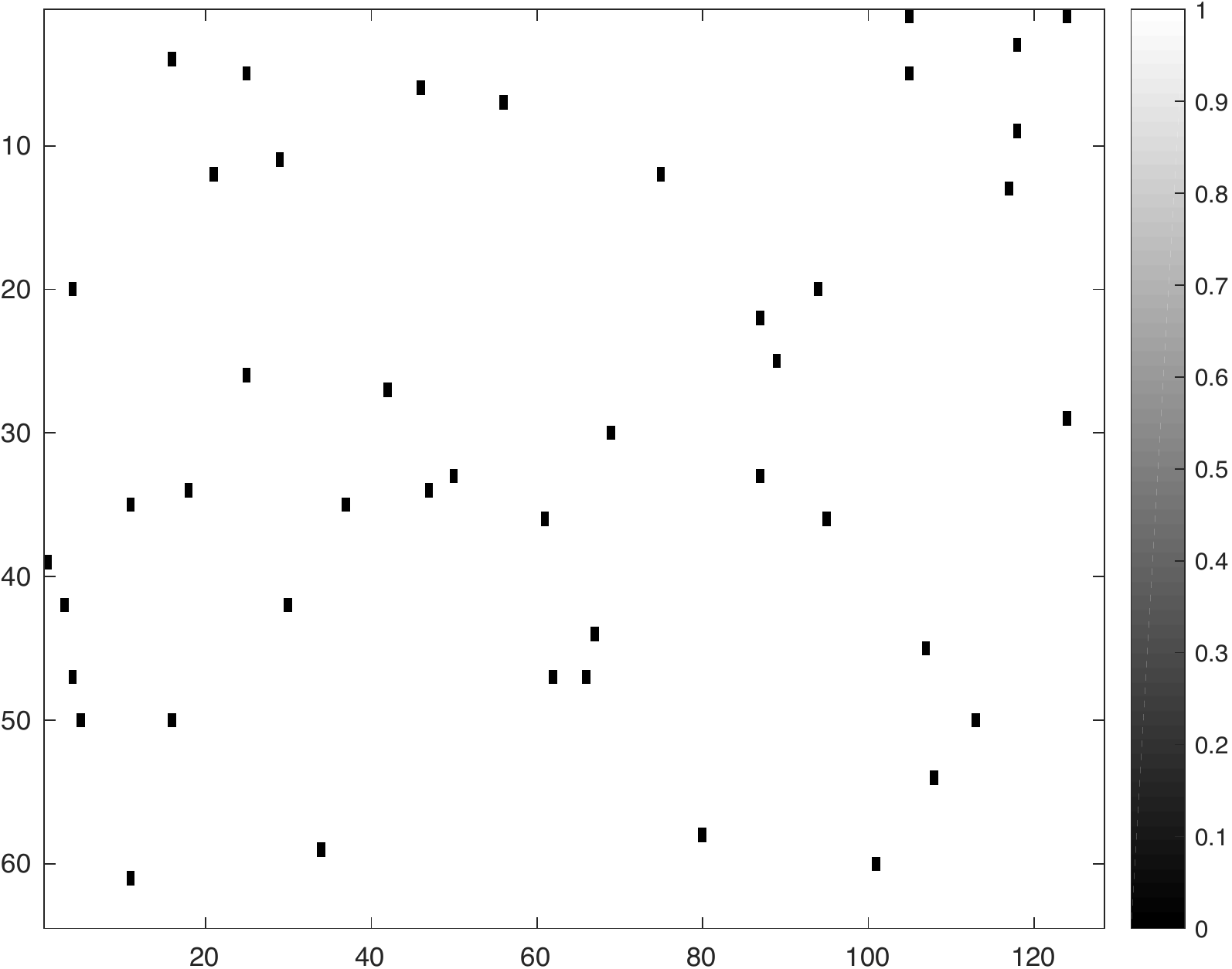}
		\caption{$\vW^\text{FC3}$ (DO+P)}
		\label{fig:W1c} 
	\end{subfigure}
~	
	\begin{subfigure}[b]{0.2\textwidth}
		\centering
		\includegraphics[width=\textwidth]{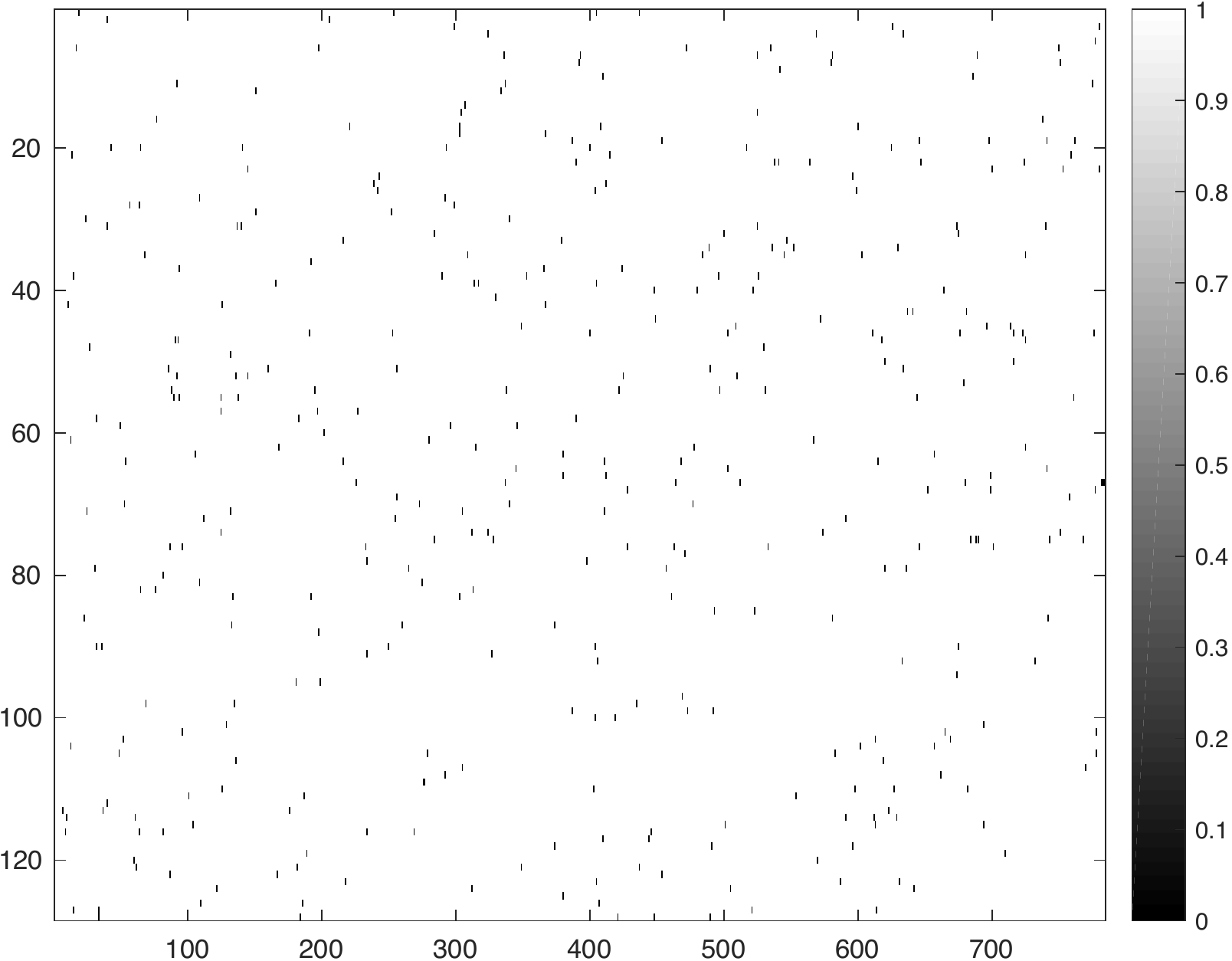}
		\caption{$\vW^\text{FC4}$ (DO+P)}
		\label{fig:W1d} 
	\end{subfigure}

	\begin{subfigure}[b]{0.2\textwidth}
		\centering
		\includegraphics[width=\textwidth]{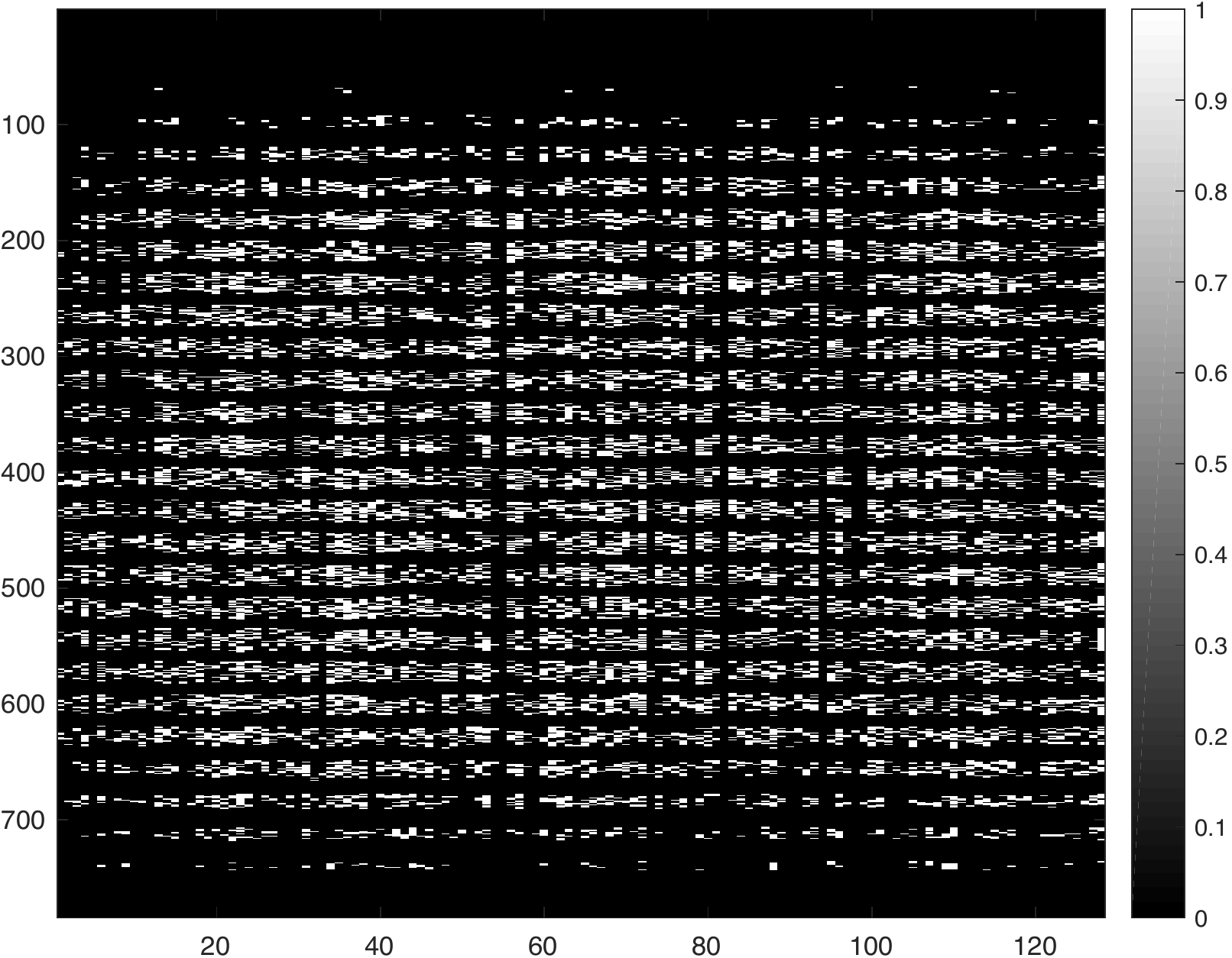}
		\caption{$\vW^\text{FC1}$ ($\ell_1$+DO+P)}
		\label{fig:W1e} 
	\end{subfigure}
~	
	\begin{subfigure}[b]{0.2\textwidth}
		\centering
		\includegraphics[width=\textwidth]{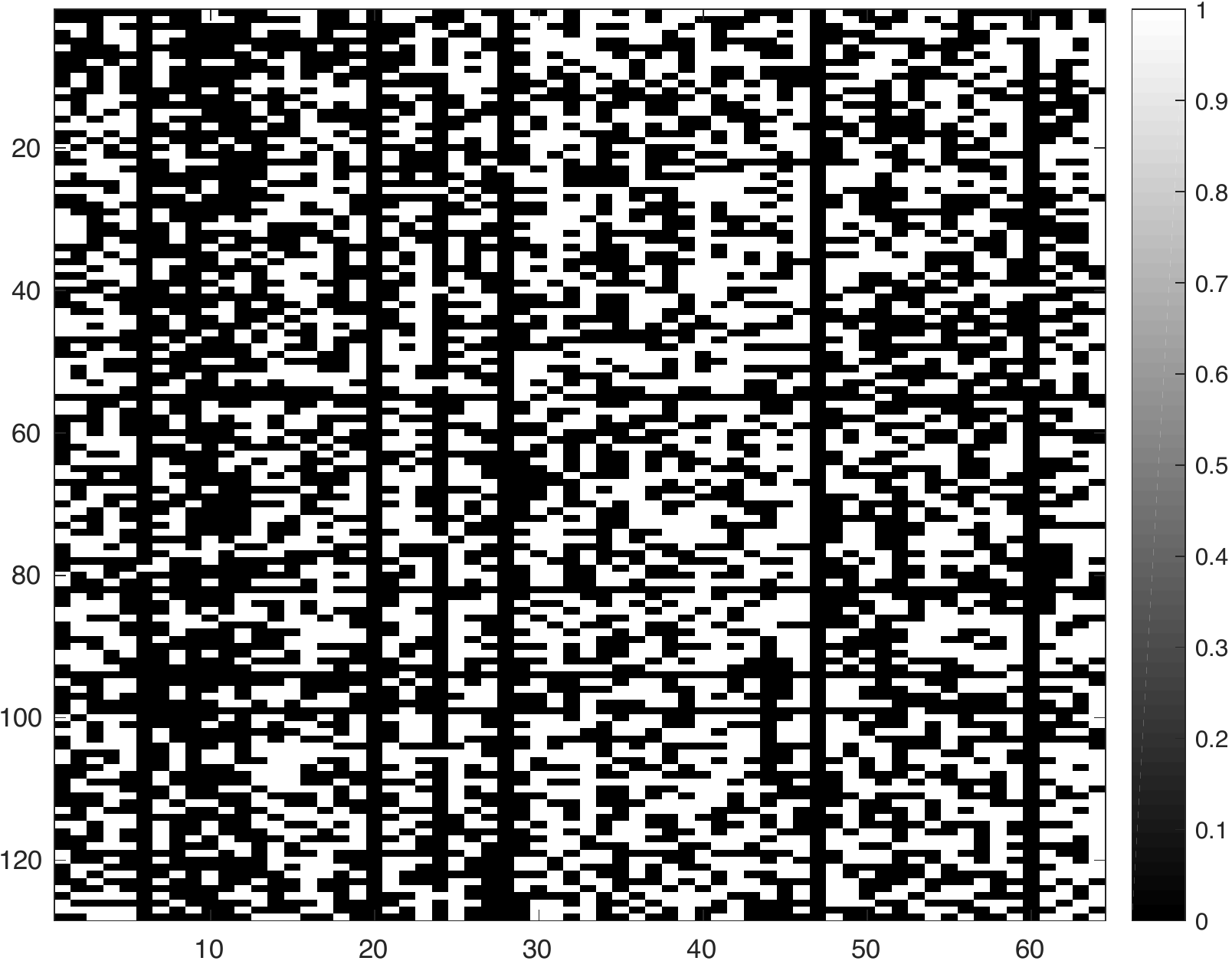}
		\caption{$\vW^\text{FC2}$ ($\ell_1$+DO+P)}
		\label{fig:W1f} 
	\end{subfigure}
~	
	\begin{subfigure}[b]{0.2\textwidth}
		\centering
		\includegraphics[width=\textwidth]{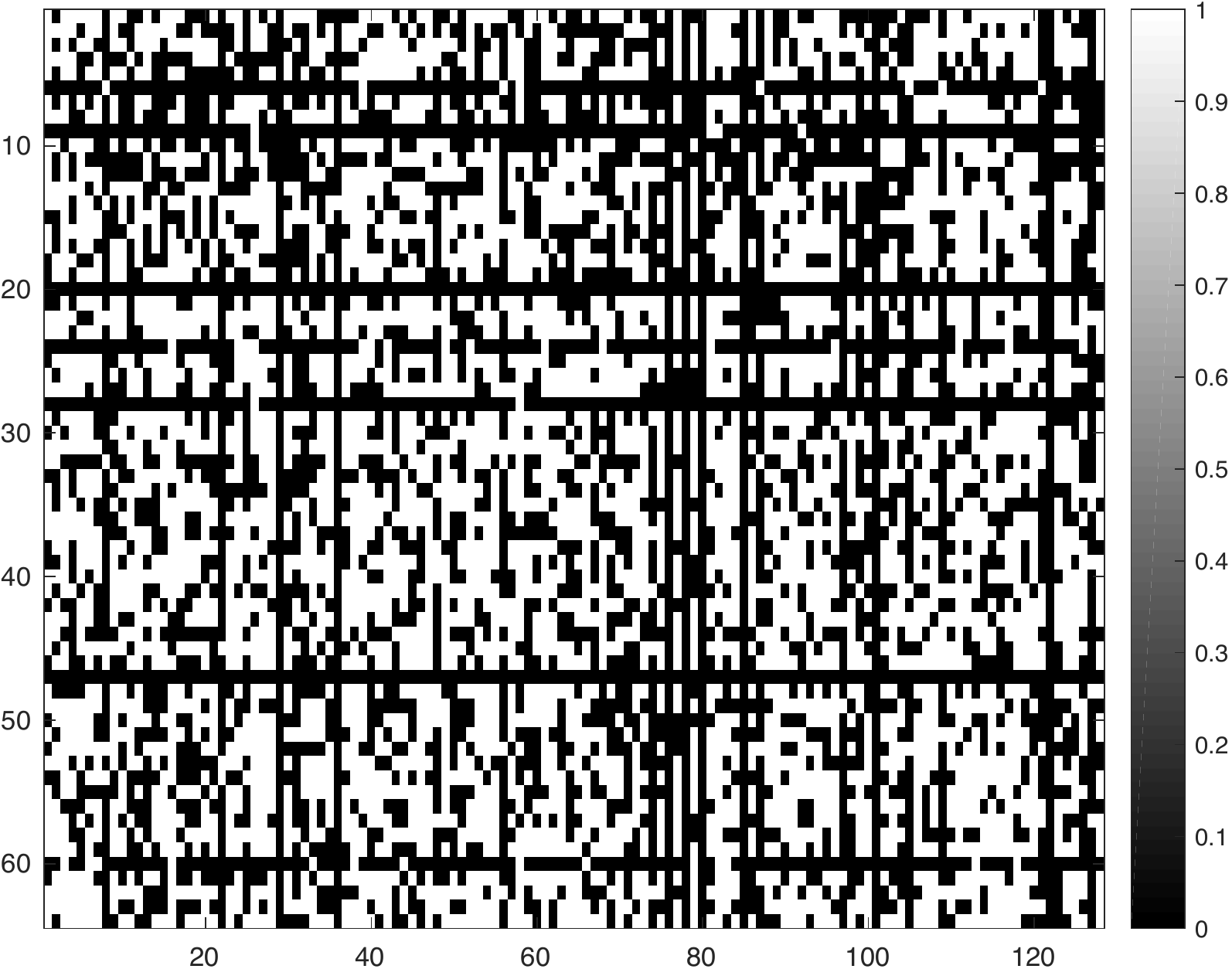}
		\caption{$\vW^\text{FC3}$ ($\ell_1$+DO+P)}
		\label{fig:W1g} 
	\end{subfigure}
~	
	\begin{subfigure}[b]{0.2\textwidth}
		\centering
		\includegraphics[width=\textwidth]{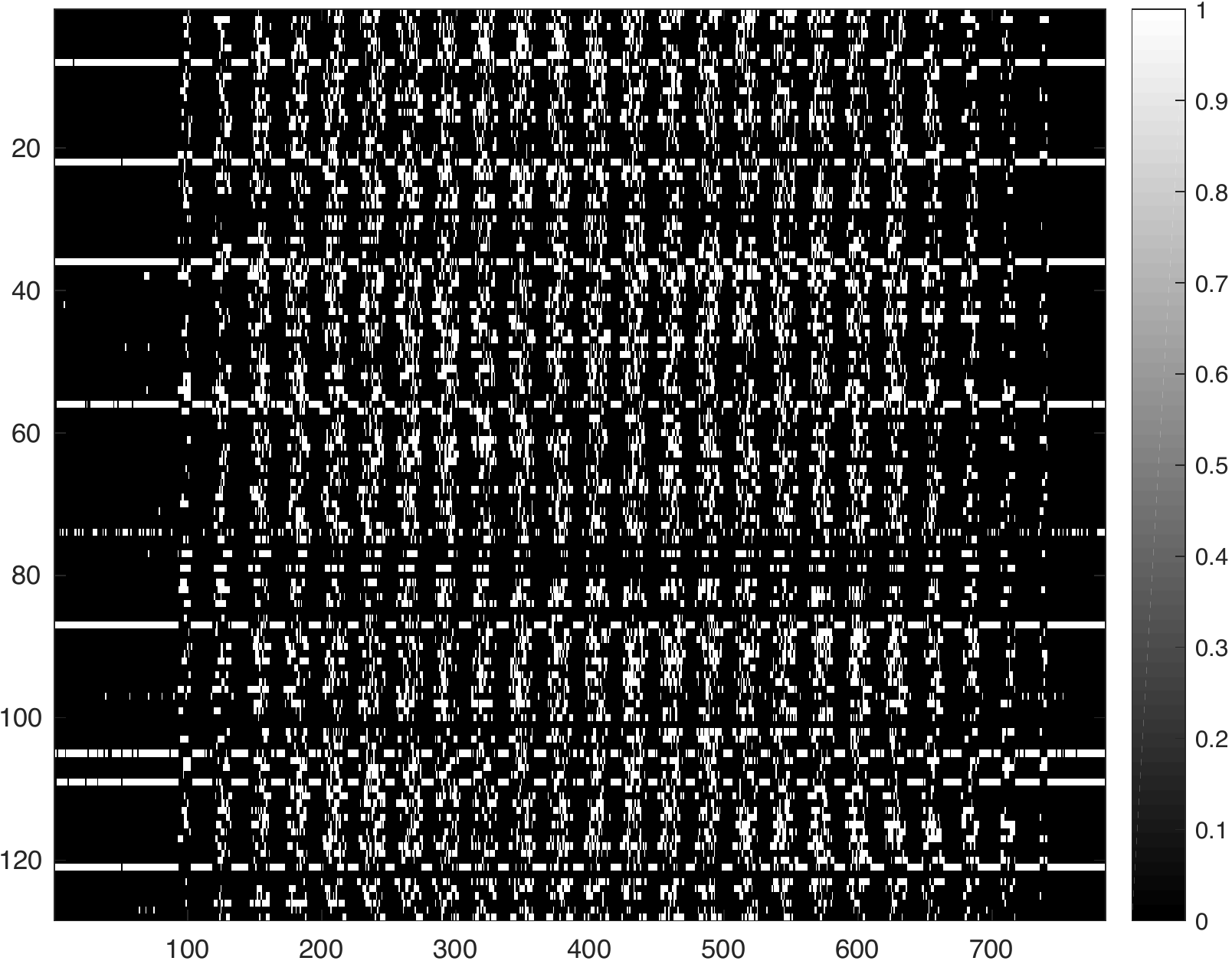}
		\caption{$\vW^\text{FC4}$ ($\ell_1$+DO+P)}
		\label{fig:W1h} 
	\end{subfigure}

	\begin{subfigure}[b]{0.2\textwidth}
		\centering
		\includegraphics[width=\textwidth]{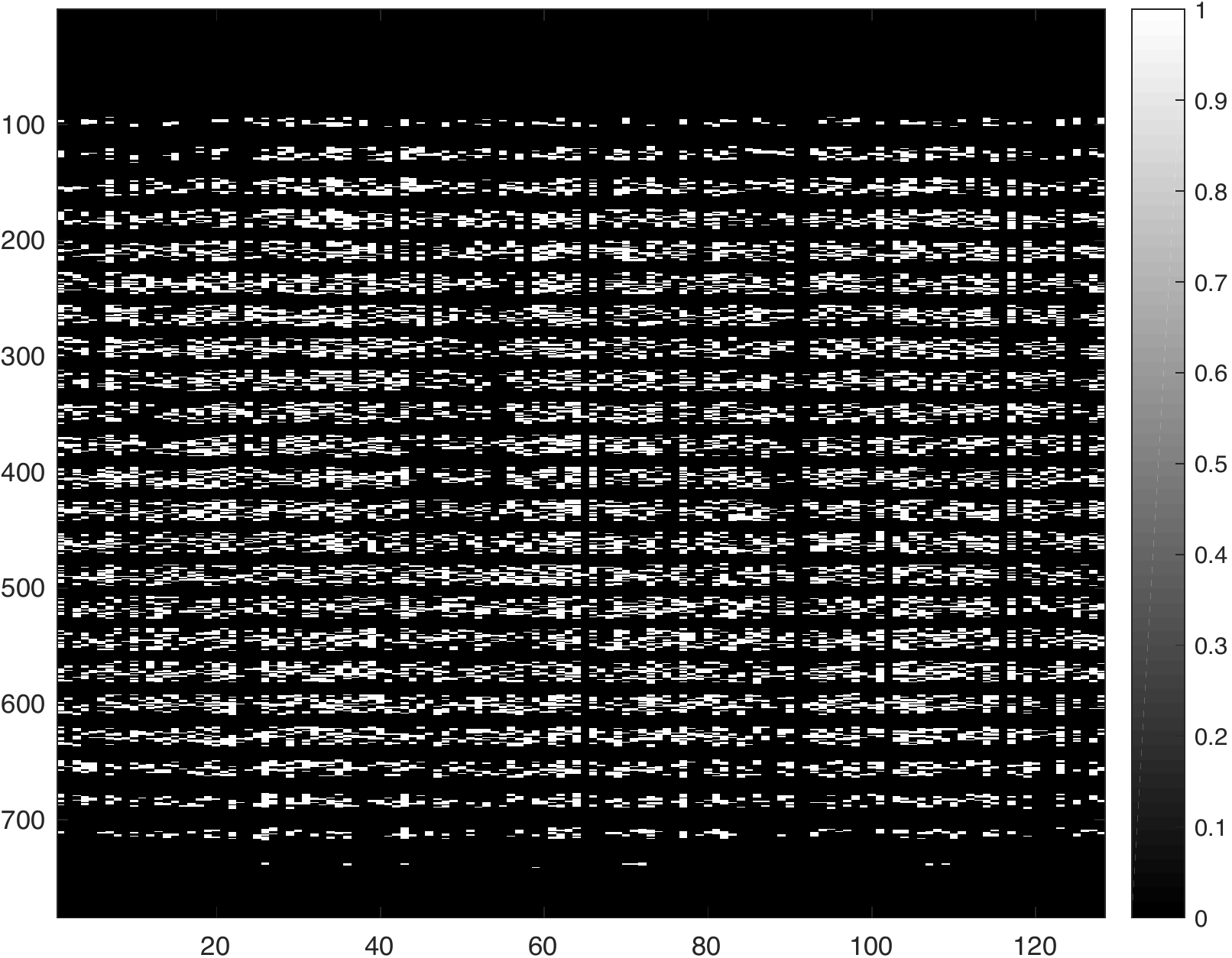}
		\caption{$\vW^\text{FC1}$ ($\ell_1$+DN+P)}
		\label{fig:W1i} 
	\end{subfigure}
~	
	\begin{subfigure}[b]{0.2\textwidth}
		\centering
		\includegraphics[width=\textwidth]{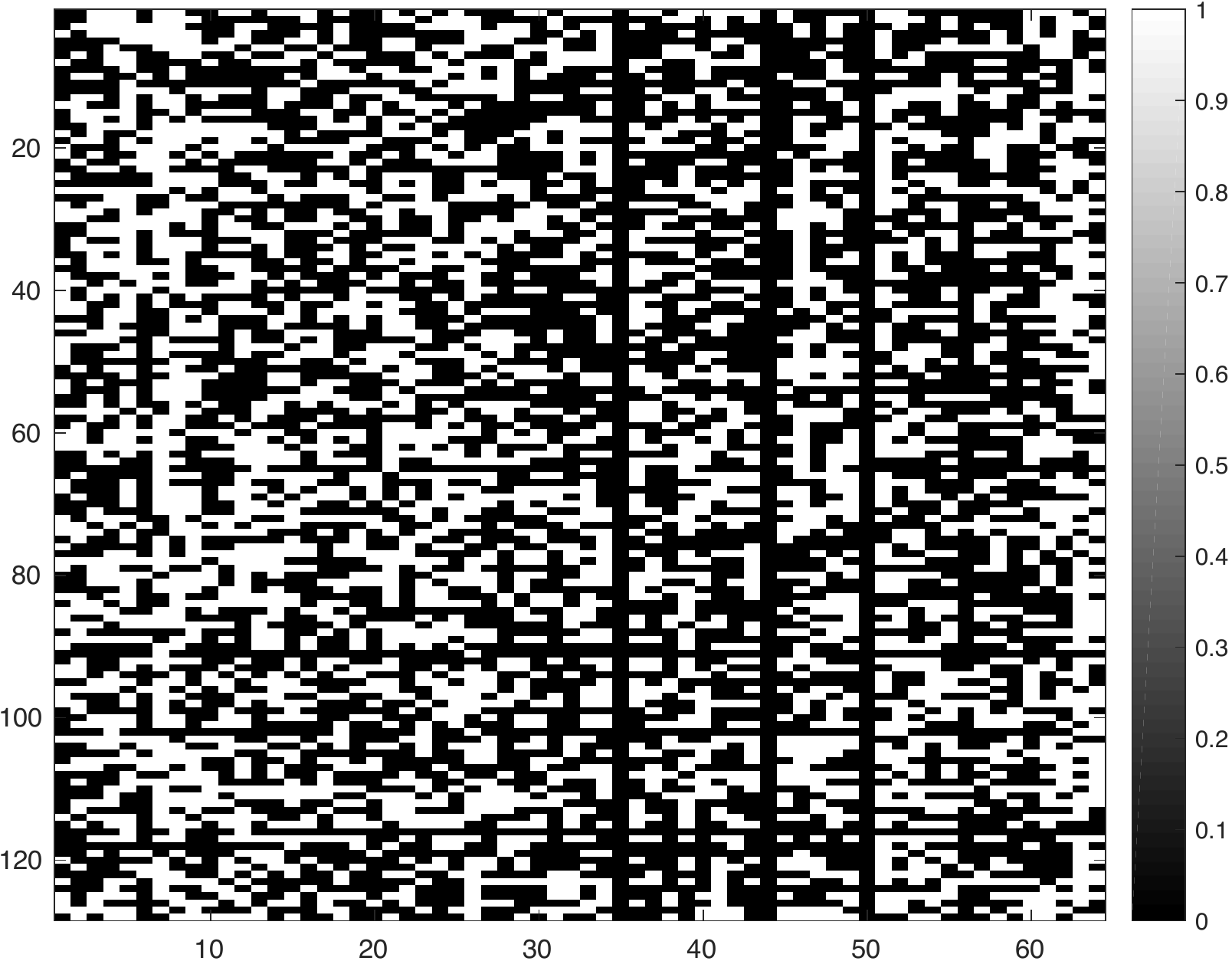}
		\caption{$\vW^\text{FC2}$ ($\ell_1$+DN+P)}
		\label{fig:W1j} 
	\end{subfigure}
~	
	\begin{subfigure}[b]{0.2\textwidth}
		\centering
		\includegraphics[width=\textwidth]{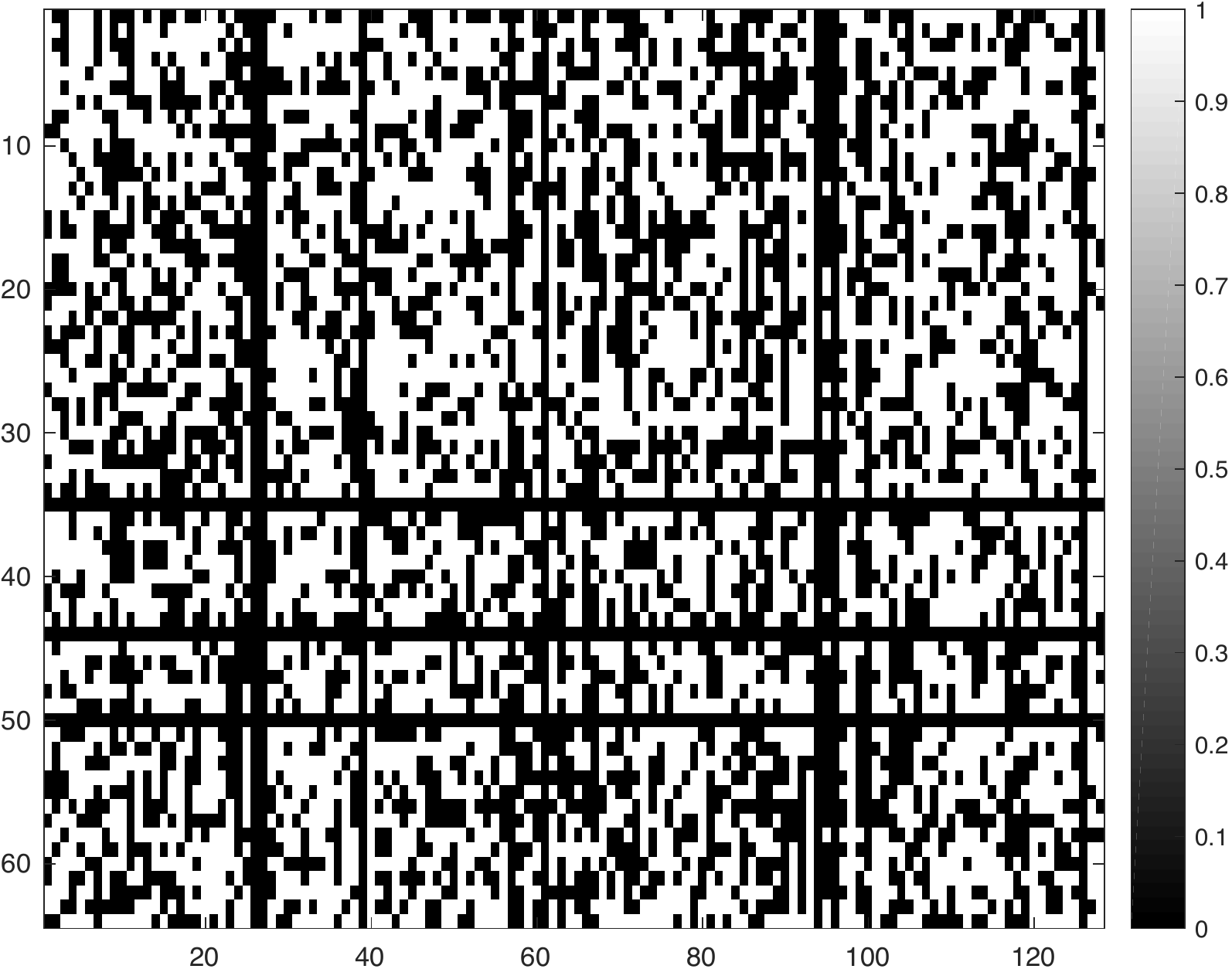}
		\caption{$\vW^\text{FC3}$ ($\ell_1$+DN+P)}
		\label{fig:W1k} 
	\end{subfigure}
~	
	\begin{subfigure}[b]{0.2\textwidth}
		\centering
		\includegraphics[width=\textwidth]{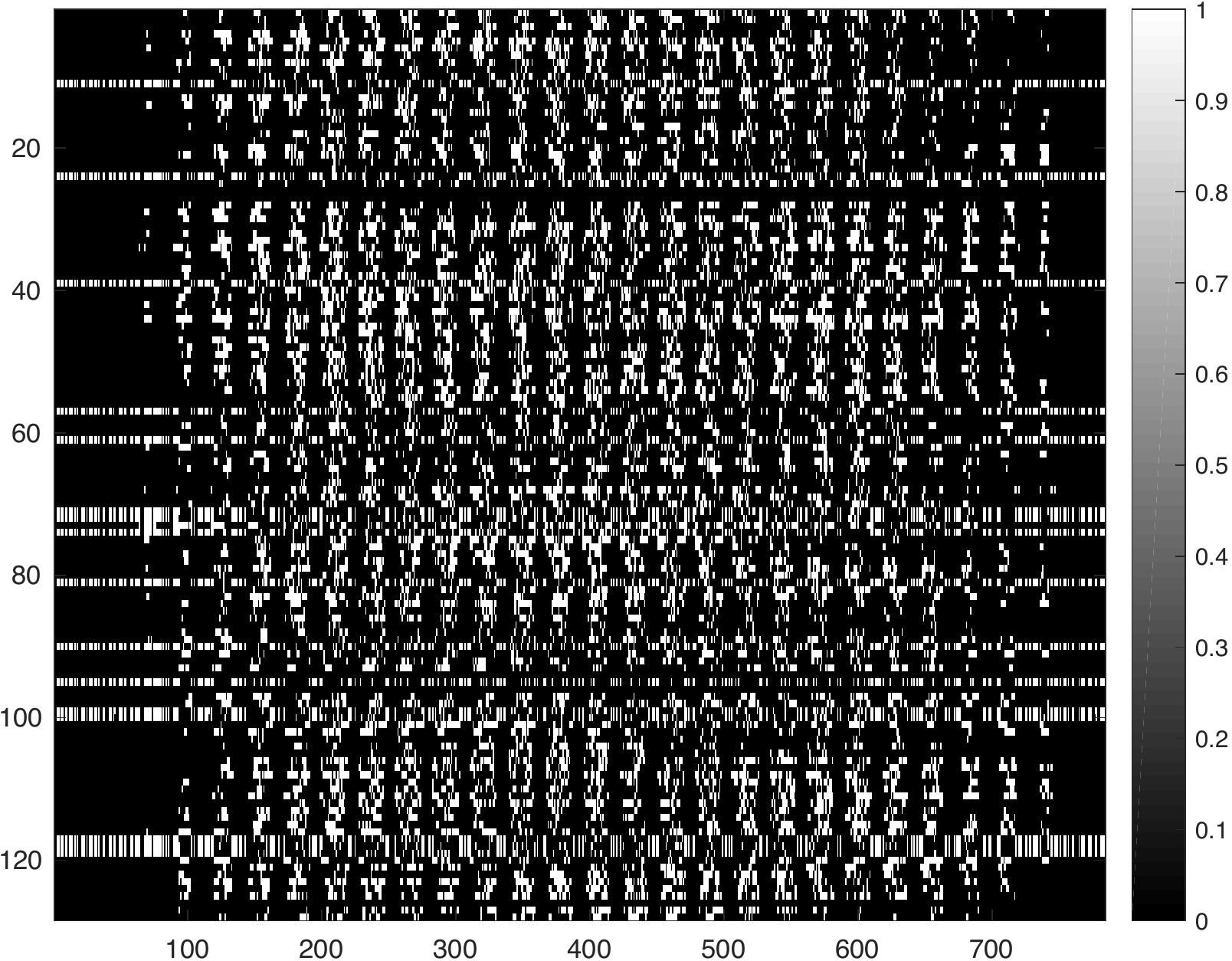}
		\caption{$\vW^\text{FC4}$ ($\ell_1$+DN+P)}
		\label{fig:W1l} 
	\end{subfigure}
	\caption{Visualisation of sparsity pattern of Autoencoder. From left to right are $\vW^\text{FC1}$ (encoder), $\vW^\text{FC2}$ (encoder), $\vW^\text{FC3}$ (decoder), $\vW^\text{FC4}$ (decoder)}
	\label{fig:W1}
\end{figure}

\begin{figure}[h]
	\centering
	\begin{subfigure}[b]{0.45\textwidth}
		\centering
		\includegraphics[width=\textwidth]{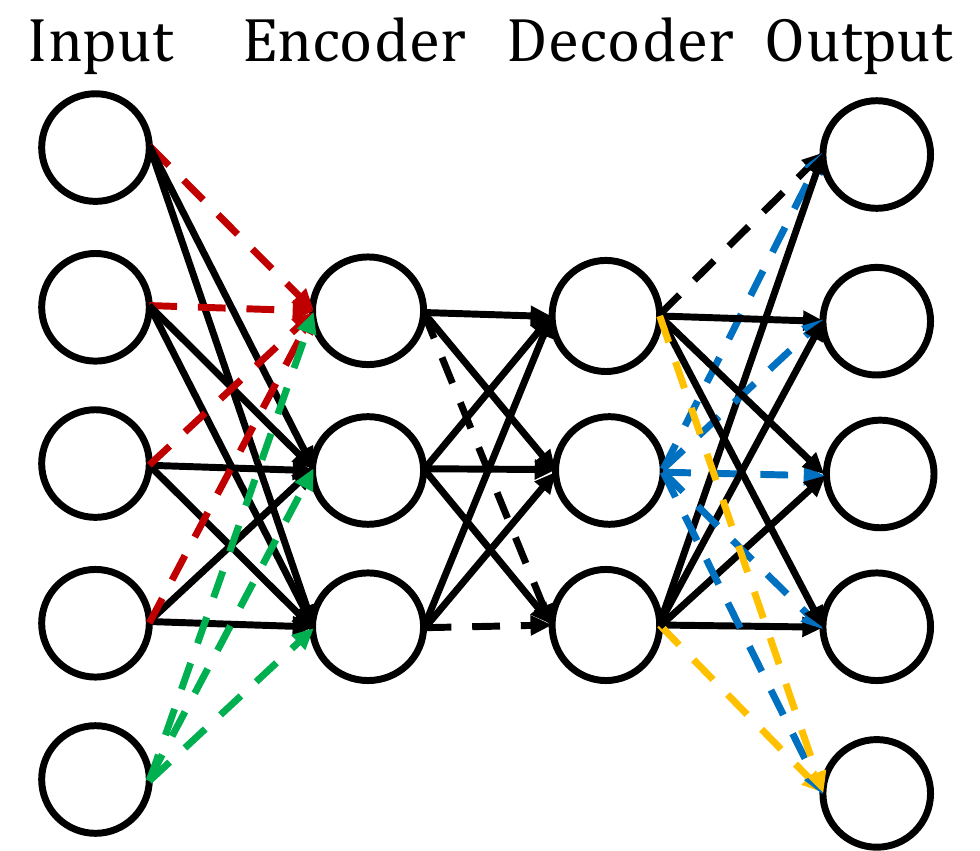}
		\caption{A fully connected autoencoder}
		\label{fig:ae1}
	\end{subfigure}
	~ %add desired spacing between images, e. g. ~, \quad, \qquad, \hfill etc. 
	\begin{subfigure}[b]{0.45\textwidth}
		\centering
		\includegraphics[width=\textwidth]{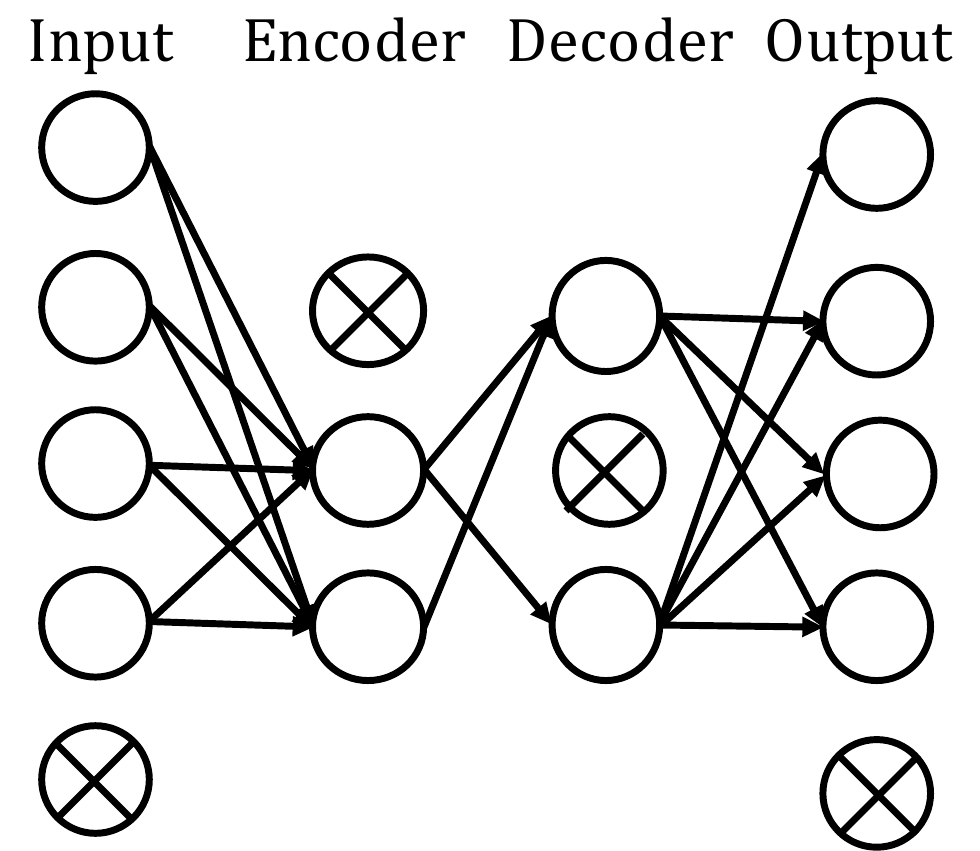}
		\caption{Drop neurons of autoencoder}
		\label{fig:ae2}
	\end{subfigure}
	\caption{The group of red dashed lines denote the regularisation of the in-coming connections to the upper neuron in the input layer.
		The group of blue dashed lines denote the regularisation of the out-going connections from the middle neuron in the output layer.
		The group of yellow dashed lines together with the lower blue dashed line denote the regularisation of the in-coming connections to the lower neuron in the output layer. 
		The black dashed lines denote the regularisation for the connections to make the alive neurons sparsely connected.
	}\label{fig:ae}
\end{figure}

\section{Example: Convolutional NN}

\begin{figure} %[h]
	\centering
	\begin{subfigure}[b]{0.4\textwidth}
		\centering
		\includegraphics[width=\textwidth]{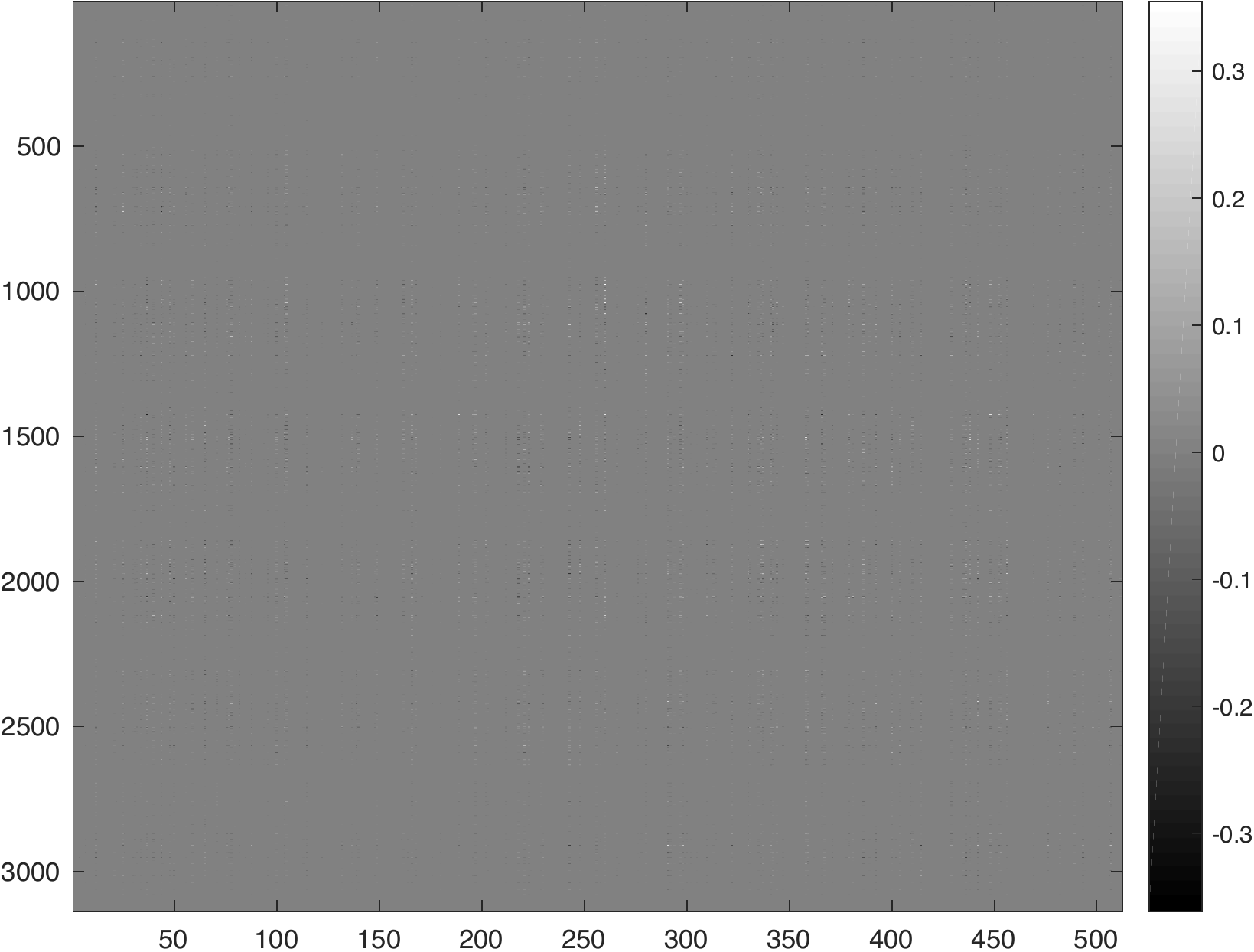}
		\caption{$\vW^\text{FC1}$ ($\ell_1$+P+DO)}
		\label{fig:1-11} 
	\end{subfigure}
~	
	\begin{subfigure}[b]{0.4\textwidth}
		\centering
		\includegraphics[width=\textwidth]{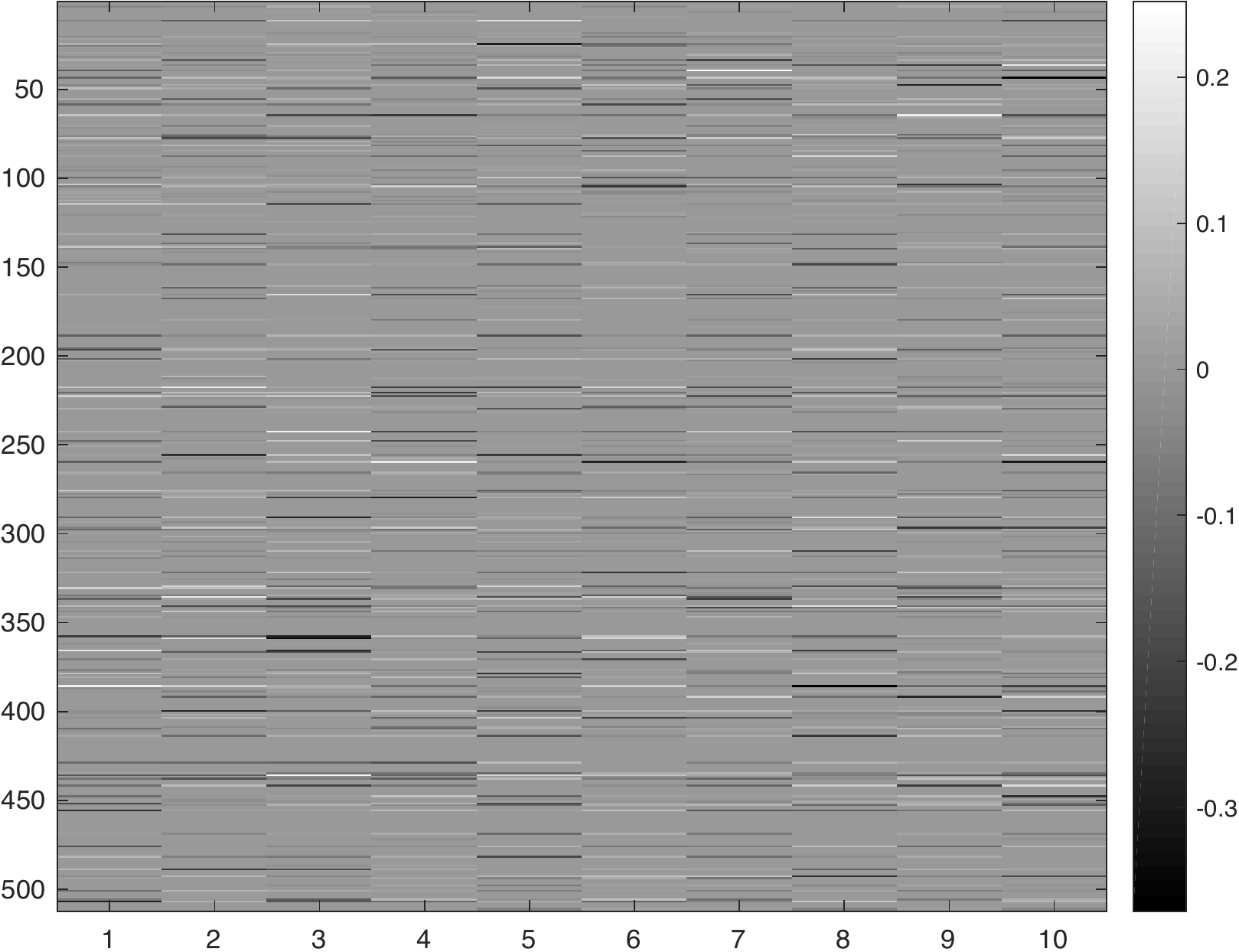}
		\caption{$\vW^\text{FC2}$ ($\ell_1$+P+DO)}
		\label{fig:1-12}  
	\end{subfigure}

	\begin{subfigure}[b]{0.4\textwidth}
		\centering
		\includegraphics[width=\textwidth]{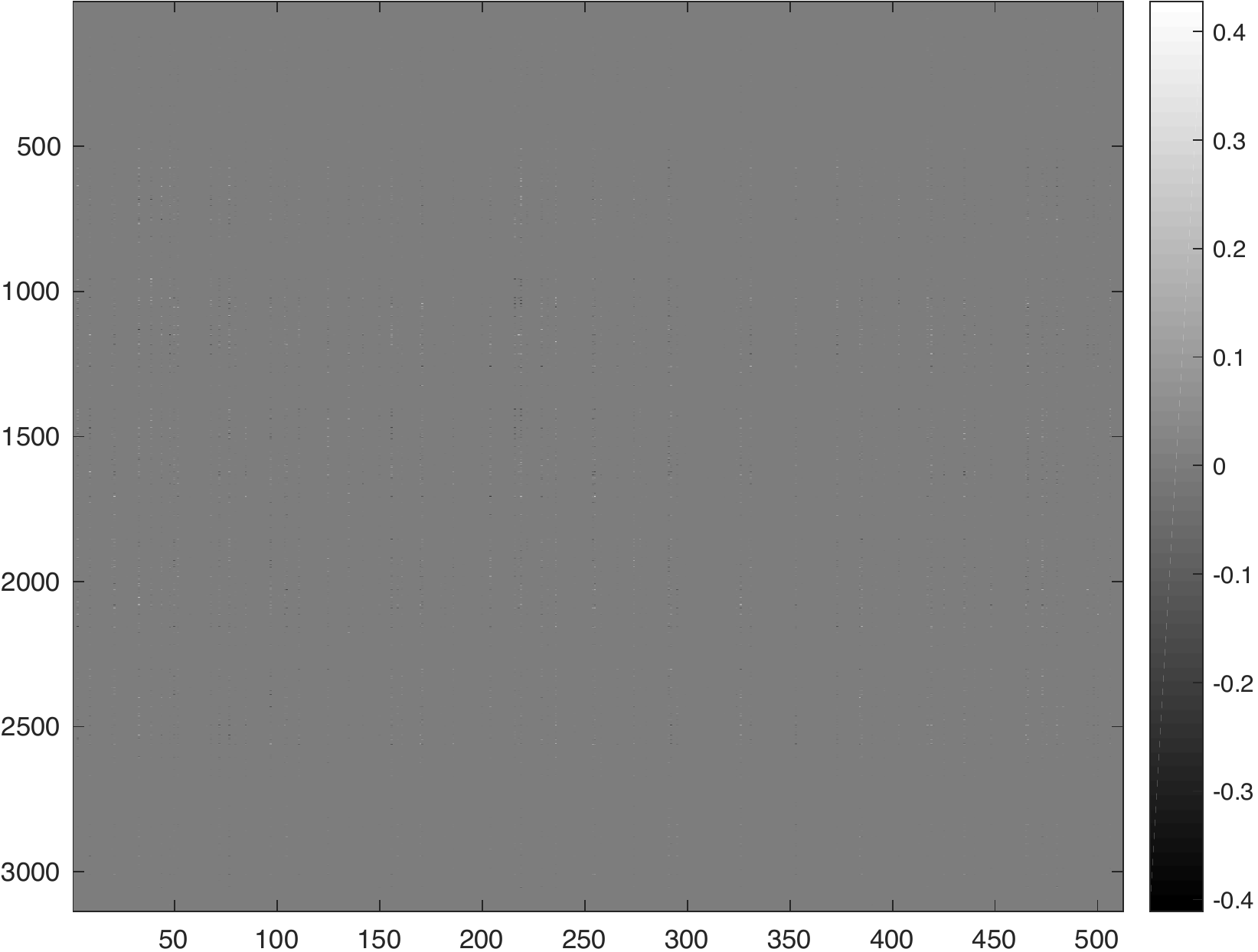}
		\caption{$\vW^\text{FC1}$ ($\ell_1$+P+DN)}
		\label{fig:1-21} 
	\end{subfigure}
~	
	\begin{subfigure}[b]{0.4\textwidth}
		\centering
		\includegraphics[width=\textwidth]{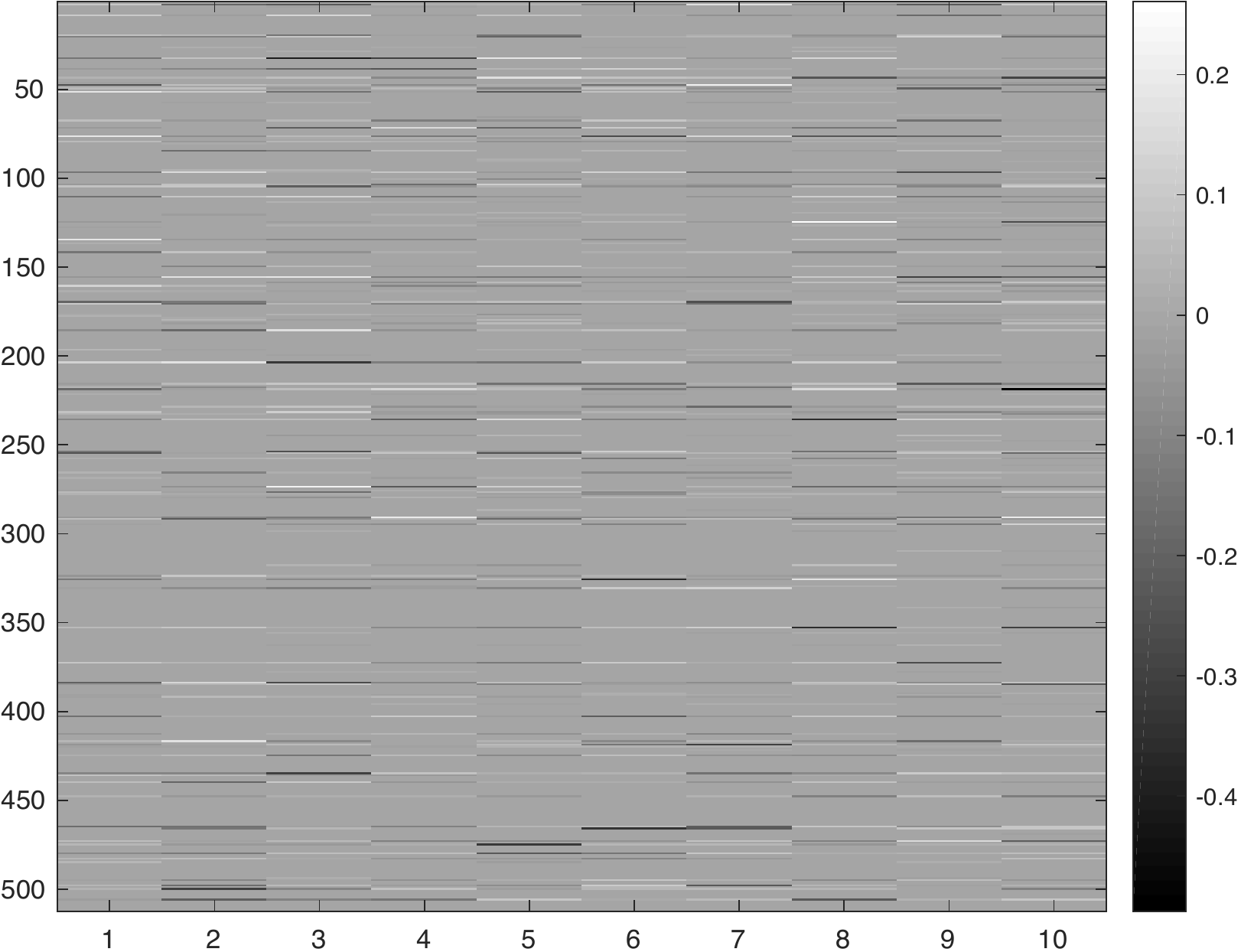}
		\caption{$\vW^\text{FC2}$ ($\ell_1$+P+DN)}
		\label{fig:1-22} 
	\end{subfigure}
		\caption{Visualisation of trained weights of FC1 and FC2  of LeNet-5 with different combination of regularisations. }\label{fig:1FC}
\end{figure}

\begin{figure} %[h]
	\centering
	\begin{subfigure}[b]{0.4\textwidth}
		\centering
		\includegraphics[width=\textwidth]{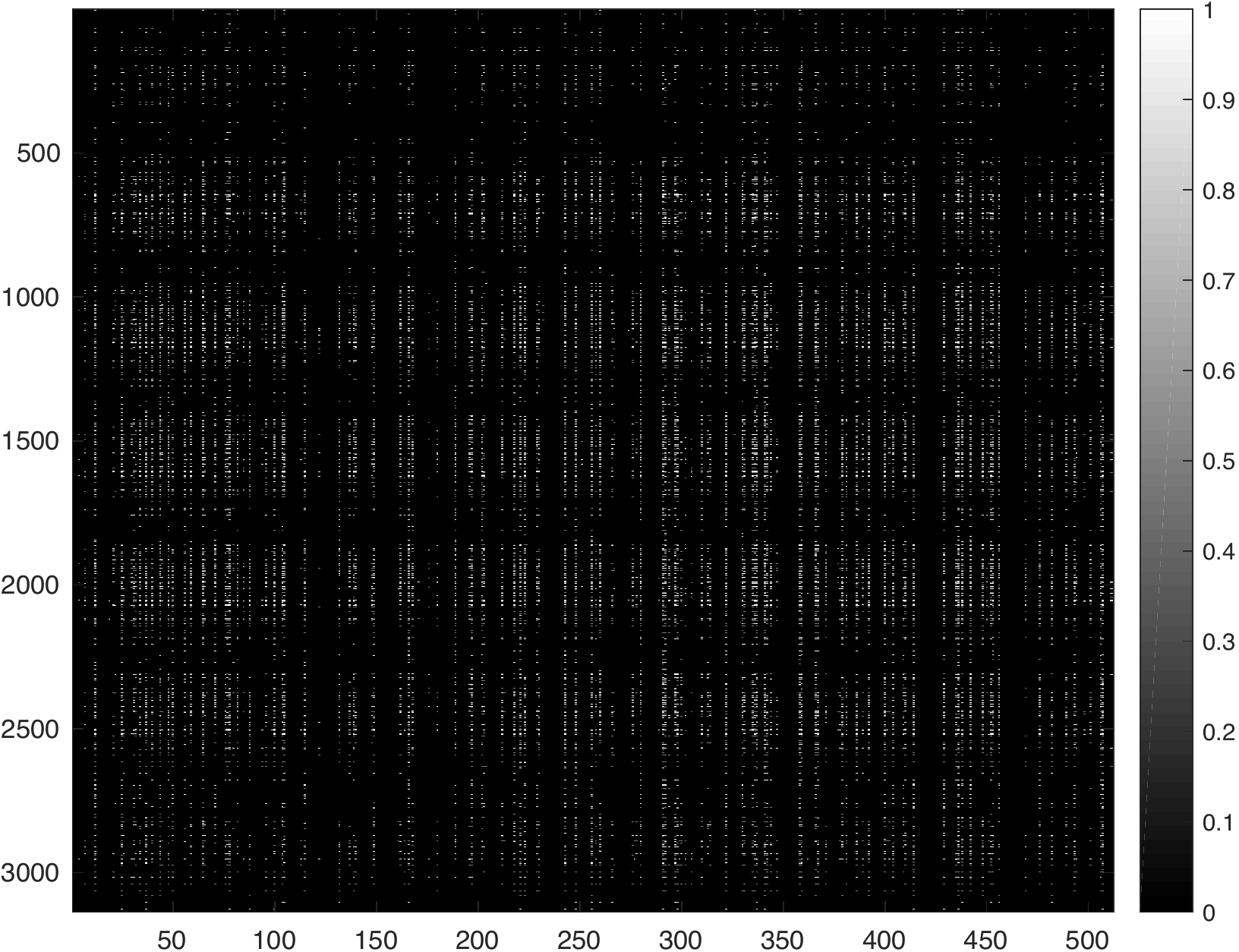}
		\caption{$\vW^\text{FC1}$ ($\ell_1$+P+DO)}
		\label{fig:2-11} 
	\end{subfigure}
~	
	\begin{subfigure}[b]{0.4\textwidth}
		\centering
		\includegraphics[width=\textwidth]{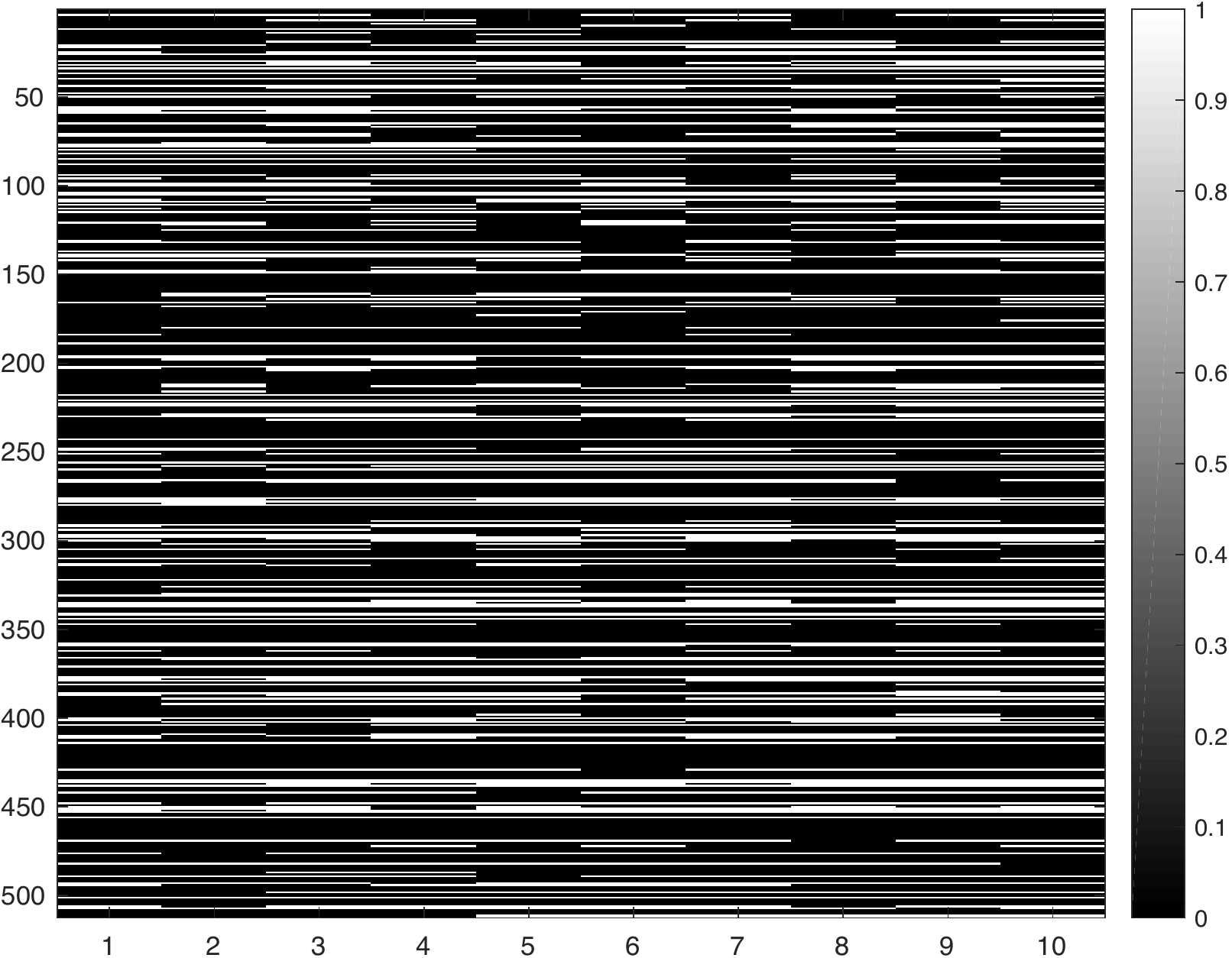}
		\caption{$\vW^\text{FC2}$ ($\ell_1$+P+DO)}
		\label{fig:2-12} 
	\end{subfigure}

	\begin{subfigure}[b]{0.4\textwidth}
		\centering
		\includegraphics[width=\textwidth]{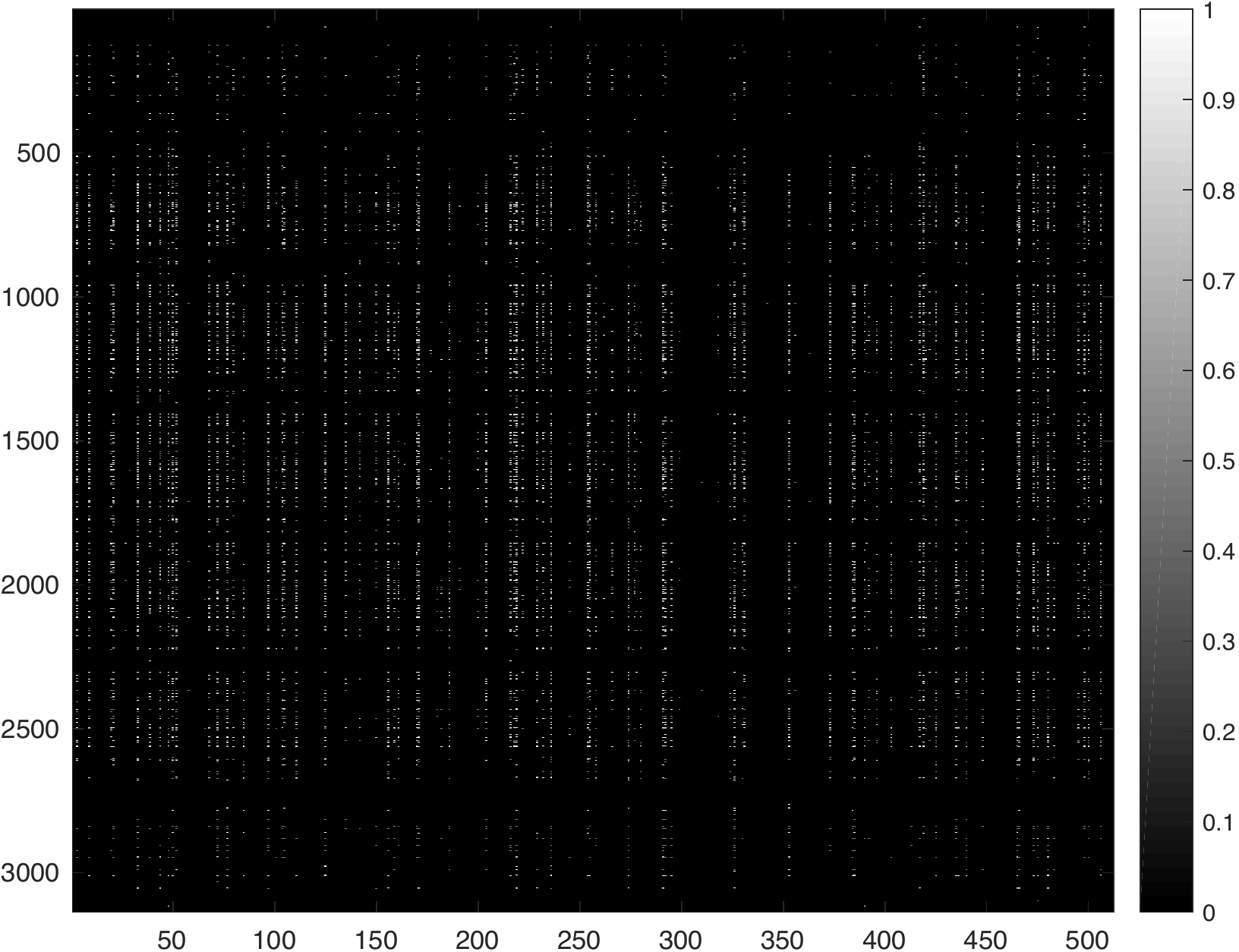}
		\caption{$\vW^\text{FC1}$ ($\ell_1$+P+DN)}
		\label{fig:2-21} 
	\end{subfigure}
~	
	\begin{subfigure}[b]{0.4\textwidth}
		\centering
		\includegraphics[width=\textwidth]{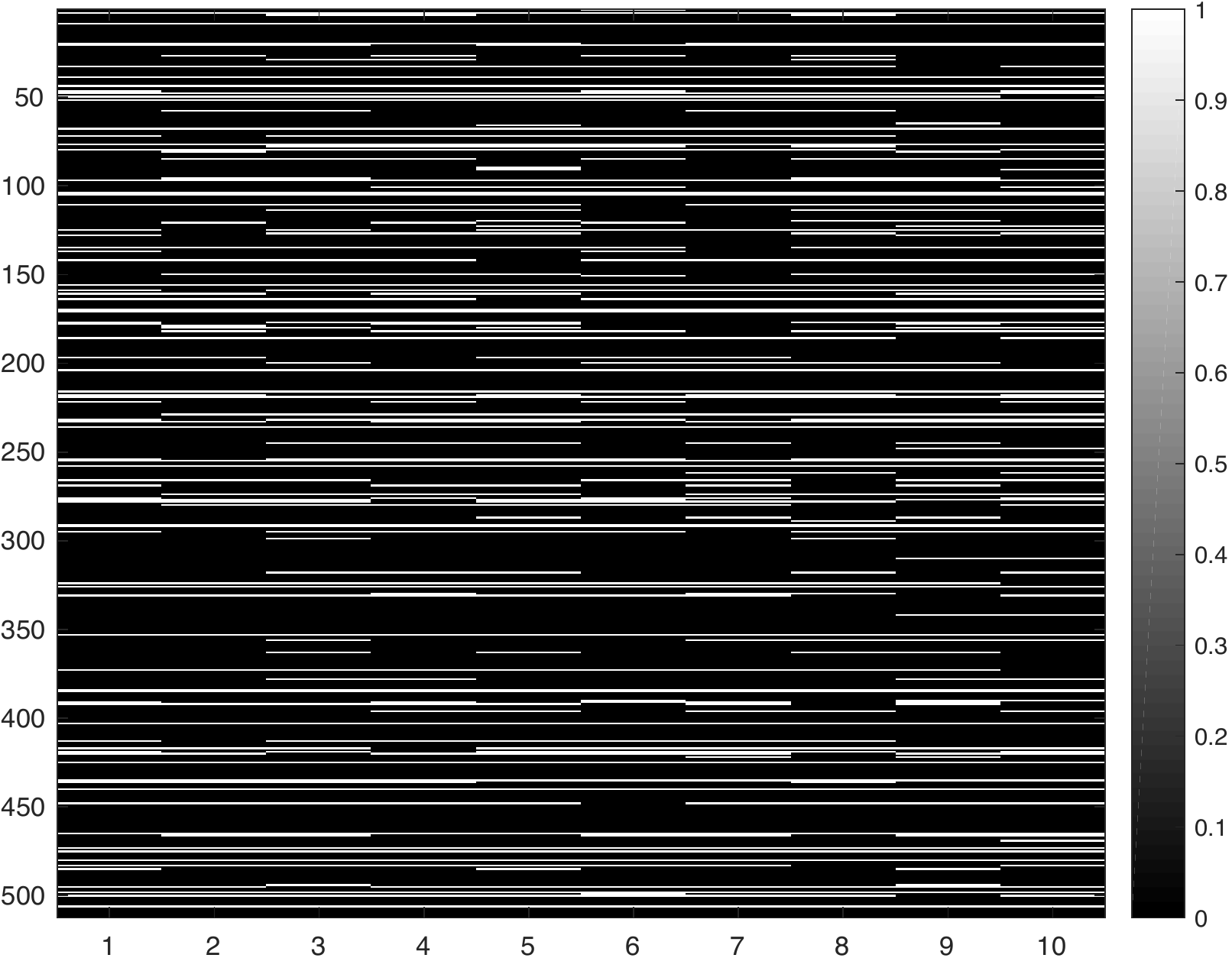}
		\caption{$\vW^\text{FC2}$ ($\ell_1$+P+DN)}
		\label{fig:2-22} 
	\end{subfigure}
	\caption{Visualisation of sparsity pattern of the FC1 and FC2 of LeNet-5 with different regularisation. All the nonzeros weights are labelled as one instead of the true value. }\label{fig:2FC}
\end{figure}

\newpage

\section*{Acknowledgements}
We acknowledge Dr Yuwei Cui for helpful discussion. We also acknowledge Dr David Birch, Mr Dave Akroyd and Mr Axel Oehmichen for setting up Linux machines with GeForce GTX 980.

\end{document}